\documentclass{article} 
\usepackage{iclr2027_conference,times}

\usepackage{amsmath,amsfonts,bm}

\def\eqref#1{equation~\ref{#1}}

\def\1{\bm{1}}

\DeclareMathAlphabet{\mathsfit}{\encodingdefault}{\sfdefault}{m}{sl}
\SetMathAlphabet{\mathsfit}{bold}{\encodingdefault}{\sfdefault}{bx}{n}

\usepackage{hyperref}
\usepackage{url}
\usepackage{graphicx}
\usepackage{booktabs}
\usepackage{multirow}
\usepackage{amsmath}
\usepackage{amssymb}
\usepackage{mathtools}
\usepackage{microtype}
\usepackage{xcolor}
\usepackage{tabularx}
\usepackage{array}
\usepackage{makecell}
\usepackage{threeparttable}
\usepackage{adjustbox}
\usepackage{placeins}
\usepackage{xspace}
\usepackage{booktabs}
\usepackage{xurl}
\usepackage{graphicx}
\usepackage{colortbl} 
\hypersetup{hidelinks}
\graphicspath{{figures/}}
\usepackage{listings}
\usepackage{float}

\newcommand{\twokm}{\ensuremath{2\,\mathrm{km}}\xspace}
\newcommand{\fivekm}{\ensuremath{5\,\mathrm{km}}\xspace}

\newcommand{\modid}[1]{\nolinkurl{#1}}

\newcolumntype{Y}{>{\raggedright\arraybackslash}X}
\newcolumntype{C}[1]{>{\centering\arraybackslash}p{#1}}
\newcolumntype{L}[1]{>{\raggedright\arraybackslash}p{#1}}

\usepackage{marvosym}   

\title{AgenticCity-Bench: Longitudinal Observations of Urban Dynamics for Multimodal Agentic Reasoning}

\author{
\textbf{Xiaofei Yu}$^{2}$\thanks{Equal contribution.
  \textsuperscript{\Letter}\,Corresponding author.} \quad
  \textbf{Kailai Sun}$^{1,3}$\footnotemark[1]    \hspace{0.2em}   \textsuperscript{\Letter} \quad
  \textbf{Yanyu Mao}$^{4}$ \quad
  \textbf{Baoshen Guo}$^{1,3}$ \quad
  \textbf{Heye Huang}$^{5}$\\
  \textbf{Yongcai Wang}$^{2}$ \quad
  \textbf{Baotong Mi}$^{2,6}$ \quad
  \textbf{Alok Prakash}$^{1,3}$ \quad
  \textbf{Shenhao Wang}$^{7}$ \quad
  \textbf{Jinhua Zhao}$^{1,3}$\\[0.5em]
  $^{1}$Singapore-MIT Alliance for Research and Technology, \hspace{0.05em}MIT \quad 
  $^{2}$Renmin University of China \quad \\
  $^{3}$Massachusetts Institute of Technology \quad 
  $^{4}$Nanyang Technological University \quad \\
  $^{5}$Korea Advanced Institute of Science and Technology\quad 
  $^{6}$Renlab AI \quad   $^{7}$University of Florida
}

\iclrfinalcopy
\begin{document}

\maketitle
\lhead{Preprint}

\begin{abstract}
City systems are shaped by interactions among infrastructure, human activity, and environmental conditions. Existing earth observation and urban benchmarks are limited in unified modeling and in capturing their joint dynamics. We introduce AgenticCity-Bench, a longitudinal multimodal benchmark for multi-layer unified state modeling and agentic city task reasoning. The benchmark covers 16 cities in 12 countries across six continents from 2000 to 2025, and includes 18 modalities spanning urban function, activity, and environmental state. It organizes heterogeneous data sources around a common spatial reference and calendar index while preserving spatial and temporal coverage, provenance and tile-level availability. On this foundation, AgenticCity-Bench supports urban cross-modal translation together with diverse agentic city tasks involving forecasting, planning and resource analysis. This study provides a unified benchmark for studying the transition from multimodal urban and earth observation to joint understanding and agentic city decision-making. The dataset is available at https://anonymous.4open.science/r/AgenticCity-Bench.
\end{abstract}

\section{Introduction}

City and earth data are essential for understanding how cities function, evolve, and respond to planning and the environment. Cities are dynamic multi-layer systems shaped by population \citep{zimmer2026globalpop}, infrastructure \citep{bettencourt2025infrastructure}, economic activity \citep{li2026satellite_ntl}, resource use \citep{li2026facilitating}, and environmental conditions \citep{liao2025temperature}. Sustainable urban development requires coordinated planning that
accounts for interactions among these components \citep{kaufmann2024democratic}. 

Recent earth observation (EO) \citep{su2026disturbance} and urban benchmarks have broadened the data and tasks available for studying cities, covering both physical landscapes \citep{li2026satellite_urbanform} and urban social activity \citep{xu2025predicting_mobility,sun2026sense}. Artificial intelligence (e.g., GenAI and agentic AI) has further accelerated these developments.  This progress has increasingly moved from single-task and single-source data toward broader, multimodal, and temporal observations\citep{yap2025revealing}. EO benchmarks \citep{nedungadi2024mmearth,marsocci2025pangaea,vanetten2021spacenet7} have expanded the use of multimodal and longitudinal observations for understanding physical environments. On the other hand, urban benchmarks \citep{feng2025citybench,zhou2025urbench,liu2026citylens} have broadened the modeling of social activity, mobility, and dynamic urban processes. These advances motivate linking social and physical knowledge and data to social activity and urban dynamics. Fine-grained alignment requires considering differences in spatial coverage \citep{o2025pointillistic}, periods, and missing data across urban records.

As urban datasets and benchmarks have evolved, they have increasingly enabled the study of agent reasoning for urban applications. In parallel, agentic AI approaches in urban intelligence have extended large language models from interpreting observations toward scientific investigation and interactive planning \citep{acbEarthAgent2026,lai2026ustbench,akinboyewa2025giscopilot}. This interplay also exposes a growing need for urban data that allows agents to connect task-relevant observations, analytical models, domain knowledge, and planning procedures. A multi-layer city should encompass not only environmental observations but also infrastructure, socioeconomic activity, and resource use, reflecting the
interconnected nature of urban systems \citep{mcphearson2016urbanecology}. However, as Table \ref{tab:compare} shows, EO benchmarks tend to focus on imagery and derived physical information, while urban benchmarks emphasize transportation, buildings, economy and other records. This specialization provides valuable but fragmented views of urban systems, with limited cross-domain coverage. 
\citep{nedungadi2024mmearth,wang2024urbandatalayer,lai2026ustbench}.
 
Integrating these data remains challenging, particularly when records span different spatial and temporal scales. Long-term static physical urban morphology and visual appearance \citep{sun2026envisioning} are different from dynamic urban activities \citep{wang2024urbandatalayer,xuan2025dynamicvl}. Current agent research has begun to explore tool use and interactive planning \citep{acbEarthAgent2026,lai2026ustbench}. These considerations highlight several limitations in existing EO/urban data and agents: (1) A multi-layer urban system needs multiple sources of data covering human activities, environmental conditions, and urban functions, unlike individual EO data or urban records alone \citep{muhlematter2025urbanfusion}. (2) Urban data should provide stable long-term coverage across heterogeneous sources, enabling joint temporal modeling. (3) Current EO agents or urban agents only support limited individual data modalities \citep{zhang2024earthgpt} to complete agentic city tasks, with most efforts still centered on conventional EO or urban datasets \citep{citygpt,xiao2026geomm}. (4) Existing EO/urban agents suffer from mismatched external tool usage \citep{feng2026earth}, a lack of city task-specific reasoning, understanding to action gaps, and inappropriate spatio-temporal scales \citep{hao2026multi}.



\begin{figure}[!htb]
    \centering
    \includegraphics[width=1\linewidth]{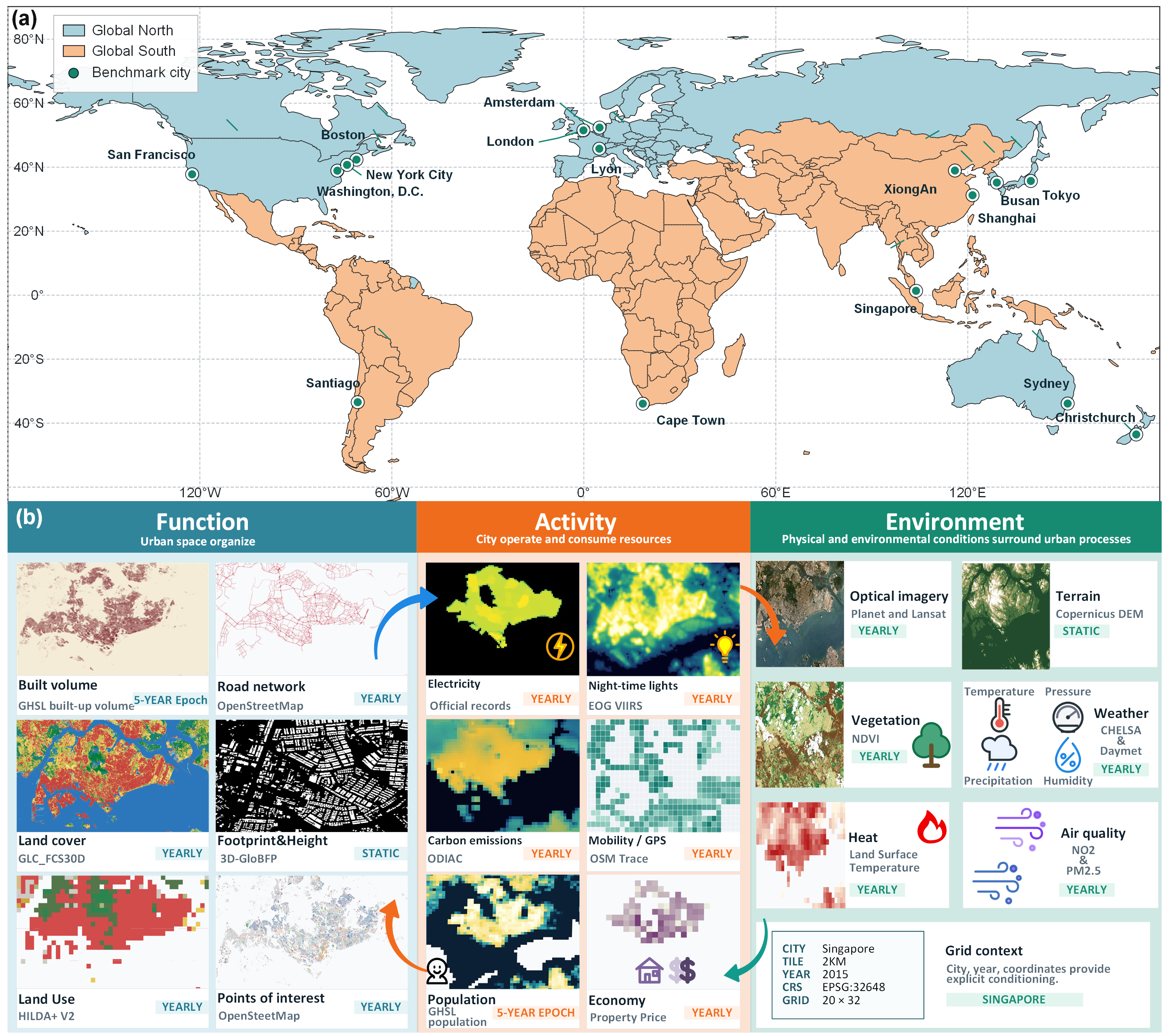}
    \vspace{-1em}
    \caption{Overview of the proposed AgenticCity-Bench. (a) Geographic distribution of the 16 cities across six continents. (b) Representative examples of the three modality groups: urban function, activity, and environment.}
    \vspace{-1em}
    \label{fig:intro}
\end{figure}

We introduce AgenticCity-Bench, a longitudinal multimodal dataset and benchmark for multi-layer system cities. It comprises 16 cities in 12 countries across six inhabited continents.  Its 18 primary modalities are divided into function, activity, and environmental information, connecting built form, infrastructure, population, property, mobility, energy, and environmental conditions through imagery, numerical and categorical grids, vector-derived data source, official records, and geo-temporal text, in Fig~\ref{fig:intro}. The 2000-2025 calendar provides a common long-term index for investigating cross-domain urban variation, while availability remains explicitly specific to each city, year, and modality. Building on this dataset, the task suite connects cross-modal state generation and analysis with urban planning and forecasting. Our contributions are as follows:
\begin{enumerate}

    \item   We propose AgenticCity-Bench, integrating 18 modalities across urban function, activity, and environmental information for 16 cities from 2000 to 2025. The dataset integrates remote sensing, infrastructure, socioeconomic activity, resource use, and environmental observations into a unified urban data framework.

    \item  An any-to-any multimodal urban translation benchmark covering 306 directed source-to-target modality pairs is established. It reveals translation relationships among all modalities, directional asymmetry, and joint training imbalance across modalities.

   \item We construct 1,000 agentic city cases spanning forecasting, urban planning, and resource analysis tasks. We evaluate diverse frontier general and domain-specific agents on multimodal urban data, analytical tools usage, and task performance. Beyond the good performance agents have achieved, experiments reveal research gaps between understanding and action, a lack of urban domain-specific knowledge, and improper tool usage.

    
\end{enumerate}
\paragraph{Paper organization.}
Section~\ref{sec:data} describes the data collection, processing,
and spatiotemporal alignment, and defines the benchmark tasks.
Section~\ref{experiments} presents the experimental setup and
results for any-to-any urban-state translation and agentic tasks.
Due to space constraints, we defer the discussion of related work
to Appendix~\ref{app:related_work}, with additional dataset and
task details provided in the subsequent Appendix~\ref{app:any2any} and Appendix~\ref{app:agentic_tasks}.

\begin{table*}[t]
\caption{Comparison with existing EO and urban datasets.
($\checkmark$) denotes a released direct semantic data counterpart;
($\triangle$) denotes a related semantic counterpart; and
-- denotes no explicitly released or documented counterpart.}
\label{tab:compare}
\centering
\scriptsize
\renewcommand{\arraystretch}{0.96}
\par\vspace{2pt}

\begin{tabularx}{\textwidth}{@{}
>{\raggedright\arraybackslash}p{0.66in}
>{\raggedright\arraybackslash}p{1.22in}
*{7}{>{\centering\arraybackslash}X}@{}}
\toprule
Family & Semantic modality
& SN7 & MME & PAN & CB & URB & CL & \textbf{Ours} \\
\midrule

\multirow{6}{0.66in}{\centering\textbf{Function}}
& Built volume (BV)
& $\triangle$ & -- & $\triangle$ & -- & -- & -- & $\checkmark$ \\
& Road network (RD)
& -- & -- & -- & $\checkmark$ & -- & -- & $\checkmark$ \\
& Land cover (LC)
& -- & $\checkmark$ & $\checkmark$ & -- & $\triangle$ & -- & $\checkmark$ \\
& Land use (LU)
& -- & -- & -- & -- & $\checkmark$ & -- & $\checkmark$ \\
& Building height \& footprint (H\&F)
& $\checkmark$ & -- & $\checkmark$ & -- & $\triangle$ & $\checkmark$ & $\checkmark$ \\
& Points of interest (POI)
& -- & -- & -- & $\checkmark$ & -- & -- & $\checkmark$ \\

\midrule

\multirow{6}{0.66in}{\centering\textbf{Activity}}
& Electricity (ELEC)
& -- & -- & -- & -- & -- & -- & $\checkmark$ \\
& Nighttime lights (NTL)
& -- & -- & -- & -- & -- & -- & $\checkmark$ \\
& Carbon emissions (CO$_2$)
& -- & -- & -- & -- & -- & -- & $\checkmark$ \\
& Mobility (GPS)
& -- & -- & -- & $\checkmark$ & -- & $\triangle$ & $\checkmark$ \\
& Population (POP)
& -- & -- & -- & $\checkmark$ & -- & $\checkmark$ & $\checkmark$ \\
& Property value (PROP)
& -- & -- & -- & -- & -- & $\checkmark$ & $\checkmark$ \\

\midrule

\multirow{6}{0.66in}{\centering\textbf{Environment}}
& Optical imagery (OPT)
& $\checkmark$ & $\checkmark$ & $\checkmark$ & $\checkmark$ & $\checkmark$ & $\checkmark$ & $\checkmark$ \\
& Terrain (DEM)
& -- & $\checkmark$ & -- & -- & -- & -- & $\checkmark$ \\
& Vegetation (NDVI)
& -- & $\triangle$ & $\triangle$ & -- & -- & -- & $\checkmark$ \\
& Weather (WTH)
& -- & $\triangle$ & -- & -- & -- & -- & $\checkmark$ \\
& Surface temperature (HEAT)
& -- & -- & -- & -- & -- & -- & $\checkmark$ \\
& Air quality (AQ: NO$_2$; PM$_{2.5}$)
& -- & -- & -- & -- & -- & -- & $\checkmark$ \\

\midrule
\multirow{1}{0.66in}{\centering\textbf{Coverage}}
& Temporal extent
& 2017--2020
& 2017--2020
& --
& --
& 2022--2023
& --
& 2000--2025 \\
\midrule

\multicolumn{2}{l}{\textbf{Matched categories (direct + related)}}
& \textbf{2+1}
& \textbf{3+2}
& \textbf{3+2}
& \textbf{5+0}
& \textbf{2+2}
& \textbf{4+1}
& \textbf{18+0} \\

\bottomrule
\end{tabularx}

\vspace{3pt}

\raggedright
\parbox{0.98\textwidth}{
\textit{Notes.}
SN7: SpaceNet~7~\citep{vanetten2021spacenet7};
MME: MMEarth~\citep{nedungadi2024mmearth};
PAN: PANGAEA~\citep{marsocci2025pangaea};
CB: CityBench~\citep{feng2025citybench};
URB: UrBench~\citep{zhou2025urbench};
CL: CityLens~\citep{liu2026citylens}.

Grid context is provided as background information during training and is not included in the 18 translation modalities.
}
\end{table*}

\begin{figure}[!htb]
    \centering
    \includegraphics[width=1\linewidth]{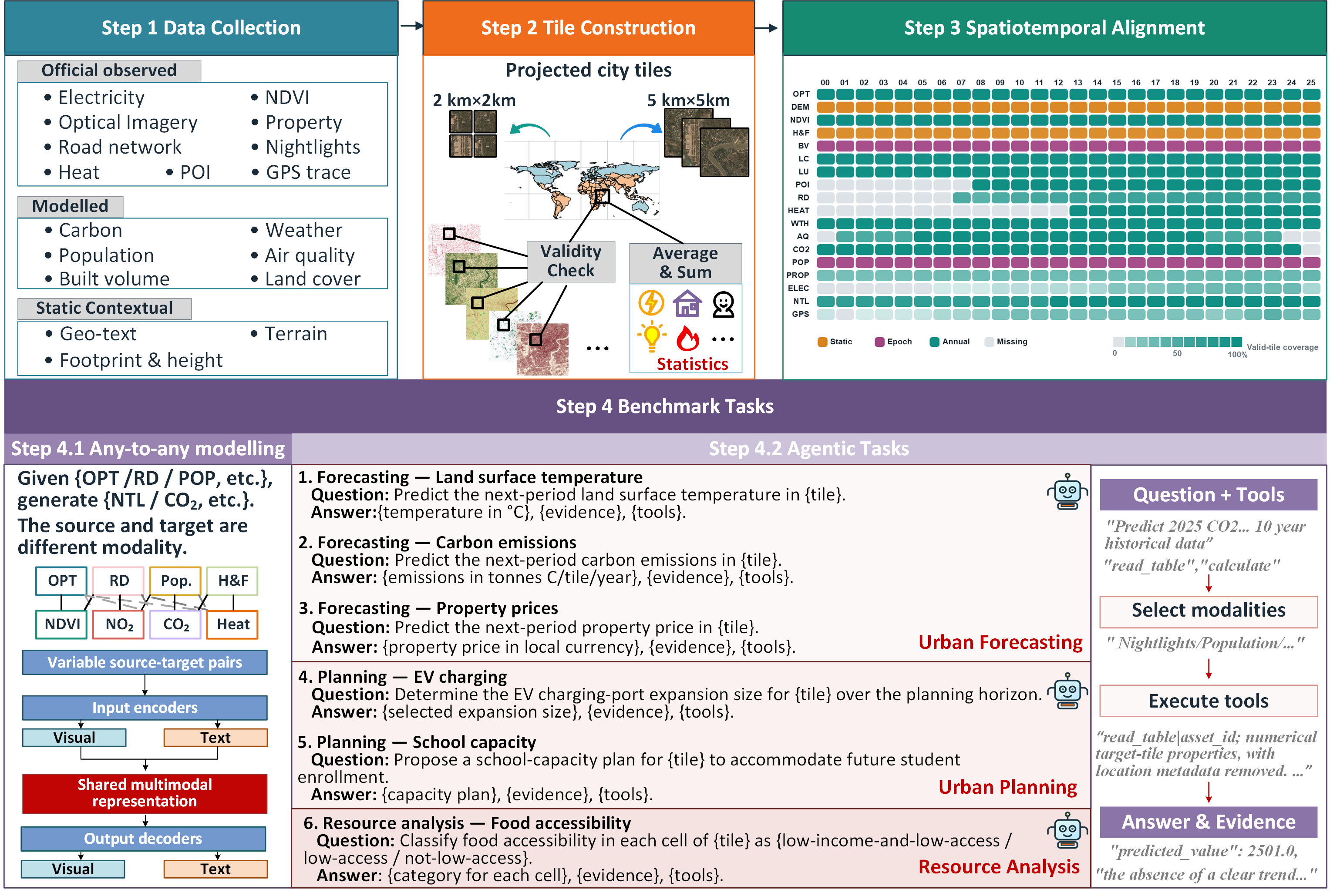}
    \vspace{-1em}
    \caption{Overview of AgenticCity-Bench. The framework comprises four steps: (1) Data Collection, which gathers multimodal urban data; (2) Tile Construction, which organizes data into standardized spatial tiles; (3) Alignment, which aligns spatial and temporal data; and (4) Benchmark Tasks, which evaluate any-to-any multimodal translation and agentic urban tasks.}
    \vspace{-1em}
    \label{fig:overview}
\end{figure}

\section{AgenticCity-Bench Establishment}
\label{sec:data}
\vspace{-1em}
AgenticCity-Bench consists of a multimodal data collection and a task framework for urban-state translation and agentic decision-making, in Fig.\ref{fig:overview}. This section describes how source data are organized into spatially and temporally indexed samples.

\subsection{Data collection.}
\vspace{-0.5em}
The modality design aims to capture complementary aspects of urban systems across space and time, supporting diverse tasks rather than a single application. We assemble 18 semantic modalities from satellite imagery, geographic databases, and official city records. The collection includes direct observations, model-derived estimates, and static conditions. Sources are selected based on their complementary information, measurement definitions, and geographic and temporal coverage.

The collection spans 16 cities in 12 countries across six continents from 2000 to 2025, with availability varying across data sources, cities, and years. In Table~\ref{tab:compare}, the 18 modalities are organized into three complementary families: (1) \textit{Function}, covering built volume, road network,  land cover, land use, and urban facilities; (2) \textit{Activity}, covering population, economic activity, mobility, electricity consumption, and emissions; and (3) \textit{Environment}, covering remote sensing, terrain, vegetation, weather, air quality, and land surface temperature. Each sample is accompanied by grid context specifying city, country, year, spatial scale, grid location, and geographic bounds. It is used only for conditioning and is not treated as a translation modality. Detailed source descriptions are provided in Appendix~\ref{app:benchmark}.

\subsection{Tile construction and processing}
\vspace{-0.5em}

\paragraph{Projected city tiles.}
Each grid divides the study area into regular, non-overlapping tiles at either $2\times2$ km or $5\times5$ km, and each tile is identified by its city, scale, row, and column. Samples are indexed by the corresponding tile and year, providing a common spatial footprint for aligning heterogeneous urban observations. Each source is then linked to the corresponding tile and year according to its spatial and temporal coverage. Annual data are aligned with the corresponding year. Five-year epoch data use the first year of each epoch for all years within that epoch \citep{tesselaar2023flood,yu2022forest}. Static data are reused across all years \citep{helfenstein2024three}. For explicitly missing intermediate years outside an epoch, no interpolation, reuse, or synthetic values are introduced; the corresponding observations are marked as missing.

\paragraph{Validity check.}
Data availability is determined separately for each modality and tile. Across all city-year-tile instances and modalities, approximately 84.255\% of observations are available. Missing or invalid observations are retained as unavailable rather than imputed. Details about data availability and quality control are provided in Appendix \ref{app:processing_alignment}.

\paragraph{Visual and statistical data.} For visual data, optical images use RGB bands without per-tile contrast adjustment. Roads, land cover, and other categorical maps use fixed colour schemes. Category labels remain discrete and are not averaged. Numerical inputs summarize valid cells within each tile. Radiance, temperature, and population density use area-weighted means. Property values are converted to 2017 international dollars for consistency. For datasets that report cell totals, values are summed and divided by tile area to obtain densities. Data-specific bands, units and spatial and temporal aggregation rules are detailed in the Appendix \ref{app:processing_alignment}.

\begin{figure}[!h]
    \centering
    \includegraphics[width=1\linewidth]{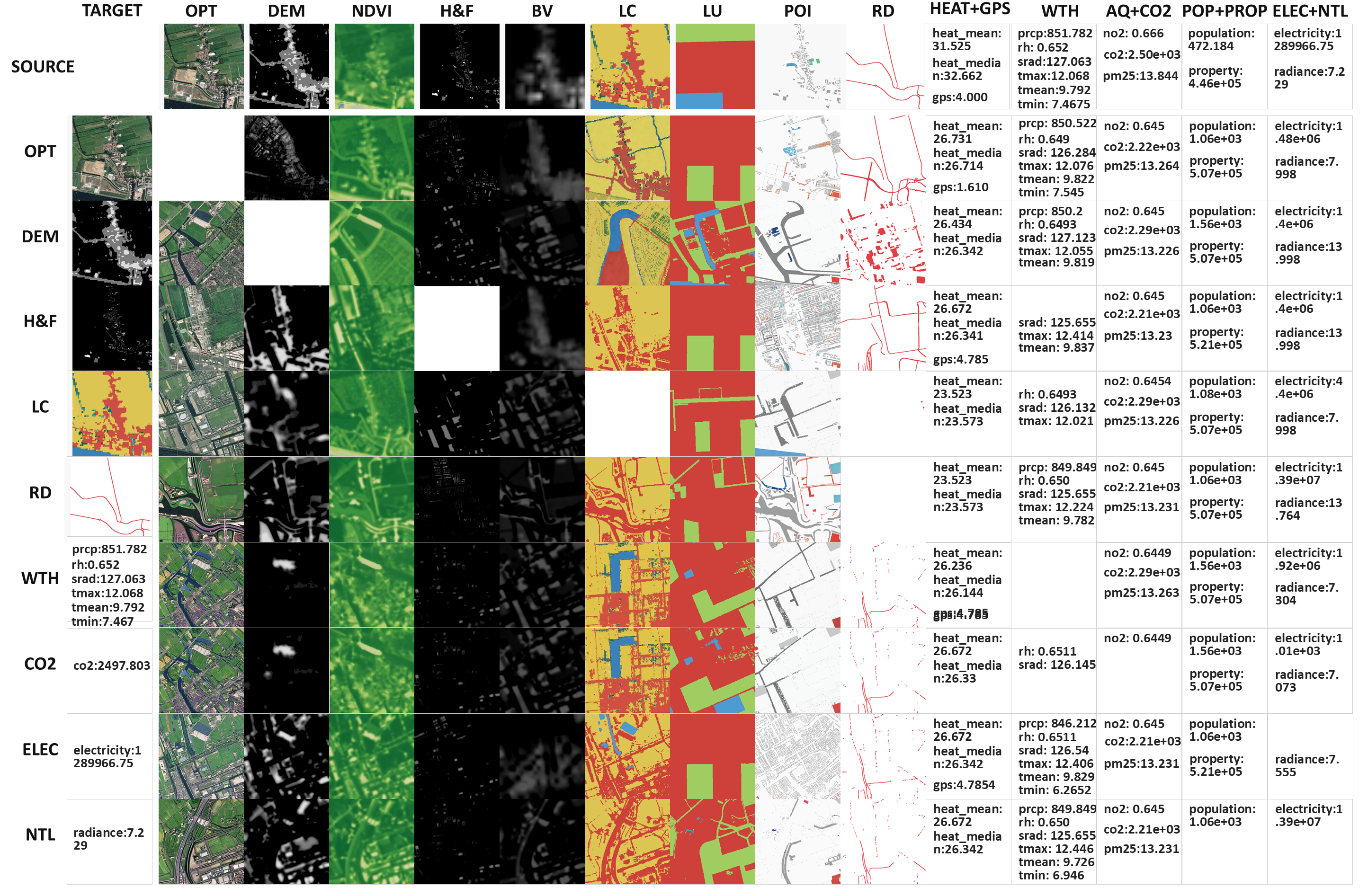}
    \vspace{-1em}
    \caption{Selected qualitative results for any-to-any cross-modal translation. Rows denote source modalities and columns denote target modalities. Same-modality pairs are omitted. The complete $18\times18$ any-2-any translation matrix is provided in Appendix~\ref{app:any2any_detailed_analysis}.}
    \vspace{-1em}
    \label{fig:half_matrix}
\end{figure}

\subsection{Benchmark Tasks }
\label{benchmark_tasks}
\vspace{-0.5em}

To systematically exploit the heterogeneous observations in AgenticCity-Bench, we construct two benchmark tracks. The first focuses on any-to-any generation across modalities. The second focuses on city agentic tasks from longitudinal records. 

\subsubsection{Any-to-Any Urban-State Translation}
\vspace{-0.5em}
AgenticCity-Bench is a unified benchmark to incorporate imagery, maps, and numerical records spanning urban function, activity, and environment, enabling bidirectional evaluation across modalities. Cross-modal translation generates a requested target modality from an available source modality. Each instance is defined within the same city, spatial scale, tile, and year, with source and target modalities $m_s,m_t \in \mathcal{M}$ and $m_s \neq m_t$, where $\mathcal{M}$ is the set of 18 modalities. The model generates the target from the source and its geographic and temporal context:
$
\hat{x}^{m_t}{c}
= f{_\theta}\left(x^{m_s}_{c}, c\right),
$
where $c$ denotes the city, spatial scale, tile, geographic coordinates, and year, and $x^{m_s}_{c}$ is the source record.

\paragraph{Training pairs and supervision.}
Source-target pairs are matched by city, tile scale, grid location, and year. Within each tile, any two valid modalities can serve as source and target in either direction. Modalities are excluded from pairing when:
(1) the modality is missing or contains invalid numerical values;
(2) Property, GPS and Electricity records must contain positive values, while valid zeros in other modalities, such as road and categorical data, are retained. In Fig.\ref{fig:overview}, for training stage, model-specific input adapters encode images or text-formatted numerical records into a shared multimodal representation. Output adapters generate the requested image or numerical record, using the corresponding target as supervision. This pairwise training setup supports different cross-modal translation directions within a unified interface.

\subsubsection{Agentic Tasks}
\vspace{-0.5em}
\label{sec:experiments}
The agentic benchmark comprises three task categories: forecasting, urban planning, and resource analysis. Each task requires the agent to plan its own workflow, select modalities, and use the retrieved information to make a prediction or spatial decision. Forecasting tasks target next land surface temperature, carbon emissions, and property prices, using historical observations indexed by relative periods $-10,\ldots,-1$ to predict period $0$. Planning tasks include EV charging port expansion and school capacity planning. The resource analysis task contains food-access analysis, which assigns each of four $1\,\mathrm{km}\times1\,\mathrm{km}$ cells within a target tile to one of three categories: low-income-and-low-access, low-access, or not-low-access. Detailed task settings are given in \ref{app:agentic-case-construction} and \ref{app:agent-prompts}.

\vspace{-0.5em}
\section{Experiments}
\label{experiments}
\vspace{-0.5em}
\subsection{Experimental Setup}

\paragraph{Any-to-any baseline}
We fine-tune and evaluate four multimodal models on AgenticCity-Bench: MODUS (7B) \cite{ye2026modus}, UniGen (Qwen2.5-1.5B backbone)\cite{tian2026unigen}, 4M-21-XL (3B) \cite{mizrahi20234m}, Show-o2 (1.5B) \cite{showo2}. All baselines are fine-tuned end-to-end from their released checkpoints. We update the full backbone, and no additional urban-specific adapter is introduced. As shown in Table \ref{tab:cooccurrence}, the dataset contains more than 28 million possible samples. We select 269,472 samples and split them into training, validation, and test sets with 172,032, 48,720, and 48,720 samples, respectively.

\paragraph{Agentic Baselines}
We evaluate frontier general agents OpenCode \citep{opencode2026software} and Codex CLI \citep{openai2026codexcli}, alongside domain-specific Earth Agent \citep{feng2026earth}. The task model configurations use GPT-5.6 Luna and GPT-5.6 Terra \citep{openai2026luna,openai2026terra}, GPT-5.5 \citep{openai2026gpt55}, DeepSeek-V4.1-Flash \citep{deepseek2026v41flash,guo2025deepseekr1}, and Gemini 3.1 Flash-Lite \citep{google2026gemini31flashlite}; within each task, configurations share the selected cases and allowed observation scope. The EV charger, school capacity planning, and food access experiments expose the same task-specific tools across their respective agent configurations, and the forecasting experiments compare Earth Agent's analytical tools with the native tools of OpenCode and Codex. 

\paragraph{Agentic task cases.}
Our agentic task comprises 1,000 cases spanning forecasting, urban planning, and resource analysis. Forecasting includes 220 land surface temperature, 220 carbon emission, and 160 property-price cases across 15, 16, and 8 cities, respectively. These cover 120 tile--target pairs across 106 distinct $2\,\mathrm{km}\times2\,\mathrm{km}$ tiles, with annual forecast origins in 2020--2024 and rolling ten-year observation windows spanning 2010--2023. Planning includes 150 EV charger location cases and 100 school capacity cases. EV cases cover 50 tiles each in Boston, New York City, and San Francisco, predicting net charging port growth from September 2024 to September 2026. School cases cover 100 London schools in distinct tiles (60 primary and 40 secondary), using enrollment and capacity histories from academic years 2021/22--2023/24 to plan additional seats for 2024/25. The sample contains 50 balanced and 50 shortage cases under target-year utilization without expansion. Resource analysis comprises 150 food access cases using 2019 reference labels: 38 each in Boston and Washington, D.C., and 37 each in New York City and San Francisco. Each tile contains four $1\,\mathrm{km}\times1\,\mathrm{km}$ cells, yielding 600 cell-level assessments.

\paragraph{Agent configurations.}
The forecasting evaluation includes seven configurations: Earth Agent with GPT Luna and DeepSeek, Codex with GPT-5.5 and DeepSeek, and OpenCode with GPT Luna, Gemini, and GLM. The added configurations use \texttt{deepseek-flash} through Codex and \texttt{glm-5.3-flash} through OpenCode. Both use the corresponding official API, a temperature of 0.1, and a requested reasoning-effort setting of low. Each added configuration evaluates the same 600 cases, comprising 220 LST, 220 carbon-emission, and 160 property-price cases. Task instructions, native tools, and case identities follow the corresponding forecasting protocol. Available observations from periods $-10$ through $-1$ are accessible for every modality, with coverage determined by each data source. Target-period outcomes remain held out for all cases.

\subsection{Any-to-Any Cross-Modal Urban Translation}
\vspace{-1em}

\begin{table*}[t]
\caption{MODUS any-to-any translation performance across target modalities.
NMSE is normalized by the training-set 1st--99th percentile range.
NMAE and NMSE values are reported as percentages.
}
\label{tab:modus-metrics}

\centering
\scriptsize
\renewcommand{\arraystretch}{1.05}
\setlength{\tabcolsep}{2pt}

\begin{tabularx}{\textwidth}{@{}l*{9}{>{\centering\arraybackslash}X}@{}}
\toprule
\multicolumn{10}{c}{\textbf{(a) Visual target modalities}} \\
\midrule
& \multicolumn{2}{c}{OPT} & DEM & NDVI & H\&F & BV & LC & LU & RD \\
\cmidrule(lr){2-3}
Input
& MSE$\downarrow$
& LPIPS$\downarrow$
& \shortstack{NMAE\\(\%)$\downarrow$}
& \shortstack{NMAE\\(\%)$\downarrow$}
& \shortstack{NMAE\\(\%)$\downarrow$}
& \shortstack{log1p NMAE\\(\%)$\downarrow$}
& \shortstack{macro\\mIoU$\uparrow$}
& \shortstack{macro\\mIoU$\uparrow$}
& clDice$\uparrow$ \\
\midrule
OPT   & --    & --    & 8.147  & 13.009 & 24.976 & 33.548 & 0.213 & 0.580 & 0.288 \\
DEM   & 0.115 & 0.614 & --     & 8.127  & 27.111 & 36.542 & 0.187 & 0.451 & 0.257 \\
NDVI  & 0.115 & 0.643 & 23.084 & --     & 28.336 & 41.056 & 0.212 & 0.507 & 0.258 \\
H\&F  & 0.120 & 0.477 & 12.681 & 12.947 & --     & 31.675 & 0.200 & 0.508 & 0.266 \\
BV    & 0.113 & 0.685 & 9.891  & 5.900  & 26.777 & --     & 0.205 & 0.452 & 0.257 \\
LC    & 0.113 & 0.671 & 26.987 & 8.217  & 21.734 & 63.800 & --    & 0.451 & 0.256 \\
LU    & 0.114 & 0.630 & 11.060 & 13.108 & 26.782 & 45.226 & 0.188 & --    & 0.257 \\
POI   & 0.117 & 0.559 & 22.600 & 8.026  & 30.799 & 43.833 & 0.199 & 0.508 & 0.260 \\
RD    & 0.109 & 0.700 & 5.982  & 9.974  & 31.358 & 54.169 & 0.196 & 0.407 & --    \\
HEAT  & 0.124 & 0.714 & 17.672 & 14.370 & 26.045 & 41.641 & 0.190 & 0.292 & 0.064 \\
WTH   & 0.126 & 0.705 & 25.116 & 8.783  & 32.715 & 69.074 & 0.111 & 0.318 & 0.096 \\
AQ    & 0.124 & 0.762 & 17.395 & 13.017 & 26.908 & 61.996 & 0.113 & 0.260 & 0.069 \\
CO$_2$ & 0.125 & 0.750 & 18.781 & 5.943  & 32.996 & 72.625 & 0.125 & 0.253 & 0.094 \\
POP   & 0.125 & 0.707 & 27.716 & 22.428 & 26.457 & 42.583 & 0.123 & 0.262 & 0.089 \\
PROP  & 0.125 & 0.733 & 24.732 & 9.206  & 28.929 & 35.753 & 0.182 & 0.368 & 0.100 \\
ELEC  & 0.124 & 0.716 & 26.461 & 8.962  & 27.102 & 38.565 & 0.169 & 0.488 & 0.077 \\
NTL   & 0.125 & 0.741 & 35.103 & 15.408 & 24.198 & 90.330 & 0.111 & 0.334 & 0.074 \\
GPS   & 0.126 & 0.738 & 29.576 & 39.560 & 26.378 & 37.917 & 0.106 & 0.267 & 0.079 \\
\bottomrule
\end{tabularx}

\vspace{5pt}

\begin{tabularx}{\textwidth}{@{}l*{9}{>{\centering\arraybackslash}X}@{}}
\toprule
\multicolumn{10}{c}{\textbf{(b) Numerical target modalities}} \\
\midrule
Input & HEAT & WTH & AQ & CO$_2$ & POP & PROP & ELEC & NTL & GPS \\
&
\shortstack{macro NMSE\\(\%)$\downarrow$}
& \shortstack{macro NMSE\\(\%)$\downarrow$}
& \shortstack{macro NMSE\\(\%)$\downarrow$}
& \shortstack{log1p NMSE\\(\%)$\downarrow$}
& \shortstack{log1p NMSE\\(\%)$\downarrow$}
& \shortstack{log1p NMSE\\(\%)$\downarrow$}
& \shortstack{log1p NMSE\\(\%)$\downarrow$}
& \shortstack{log1p NMSE\\(\%)$\downarrow$}
& \shortstack{log1p NMSE\\(\%)$\downarrow$} \\
\midrule
OPT    & 13.756 & 9.333  & 23.656 & 13.450 & 28.571 & 10.287 & 4.093  & 19.847 & 28.800 \\
DEM    & 7.537  & 0.756  & 19.861 & 9.347  & 25.000 & 14.294 & 24.822 & 42.301 & 20.015 \\
NDVI   & 8.208  & 12.397 & 32.573 & 3.196  & 33.056 & 13.134 & 28.571 & 14.764 & 28.255 \\
H\&F   & 13.694 & 16.015 & 25.851 & 8.046  & 20.229 & 14.286 & 14.286 & 12.200 & 15.504 \\
BV     & 14.340 & 2.356  & 28.102 & 19.763 & 16.262 & 6.370  & 10.994 & 12.605 & 15.366 \\
LC     & 15.693 & 1.442  & 10.153 & 1.520  & 29.078 & 11.966 & 12.500 & 22.715 & 30.037 \\
LU     & 20.715 & 9.327  & 40.779 & 7.006  & 28.571 & 6.018  & 56.890 & 33.880 & 13.584 \\
POI    & 6.484  & 1.847  & 14.737 & 30.605 & 24.785 & 11.150 & 37.500 & 19.628 & 14.605 \\
RD     & 9.270  & 1.024  & 30.054 & 17.489 & 28.207 & 19.842 & 37.500 & 30.562 & 13.720 \\
HEAT   & --     & 12.538 & 14.153 & 23.190 & 14.885 & 10.476 & 30.947 & 27.971 & 16.174 \\
WTH    & 29.719 & --     & 49.125 & 30.252 & 12.699 & 13.815 & 21.316 & 11.088 & 21.540 \\
AQ     & 18.009 & 2.128  & --     & 33.890 & 12.500 & 14.041 & 12.789 & 25.761 & 13.064 \\
CO$_2$ & 6.315  & 0.667  & 26.212 & --     & 28.945 & 17.242 & 6.425  & 14.821 & 29.400 \\
POP    & 24.948 & 4.944  & 27.970 & 1.078  & --     & 0.855  & 1.318  & 10.608 & 12.716 \\
PROP   & 15.385 & 19.973 & 34.070 & 4.751  & 48.012 & --     & 12.500 & 34.104 & 12.716 \\
ELEC   & 14.185 & 12.063 & 37.791 & 9.740  & 2.947  & 3.953  & --     & 2.450  & 8.118 \\
NTL    & 25.261 & 12.889 & 15.405 & 0.845  & 4.400  & 9.019  & 6.796  & --     & 2.488 \\
GPS    & 5.503  & 10.093 & 41.834 & 20.613 & 0.219  & 4.796  & 5.947  & 7.708  & --    \\
\bottomrule
\end{tabularx}

\end{table*}

Tables~\ref{tab:modus-metrics} and~\ref{tab:5route} evaluate translation between any two modalities and provide representative cross-model comparisons. Three main findings follow. (1) Any-to-any translation reveals cross-modal complementarity and translation relationships. This task provides a way to examine how information in one modality can support the generation of another.  For example, Table~\ref{tab:modus-metrics} shows that activity modalities such as population, electricity, nighttime lights, GPS density, and carbon emissions perform well on several numerical-to-numerical translations. Population yields better generation results for electricity than continuous visual features such as the road network. (2) Cross-modal translation is directionally asymmetric. Translation performance varies by direction. Visual inputs can predict numerical targets, but numerical inputs are less effective at recovering spatial structure. In Fig.~\ref{fig:matrix}, numerical inputs can generate plausible images and land-cover maps, but their spatial layouts may differ from the ground truth. (3) Joint any-to-any training is uneven across modalities. Although optical imagery and road maps both capture urban spatial structures, their visual characteristics differ substantially. In Fig \ref{fig:half_matrix}, the jointly trained model can generate spatially plausible optical imagery but struggles to generate coherent road networks. In Table~\ref{tab:5route}, different models perform better on different tasks rather than showing consistent performance across modalities.

In summary, these results highlight that cross-modal translation offers a way to further explore relationships among modalities, while pointing to the need for stronger modality-aware optimization and joint-training strategies.

\begin{table*}[!h]
\caption{Performance on five representative source-target urban translation routes. -- indicates that the model does not natively support image-to-image generation.}
\label{tab:5route}
\centering
\setlength{\tabcolsep}{3.0pt}
\renewcommand{\arraystretch}{1.10}
\resizebox{\textwidth}{!}{%
\begin{tabular}{lccccc}
\toprule
Models
& \makecell{NDVI $\to$ WTH\\macro NMSE (\%)$\downarrow$}
& \makecell{POI $\to$ ELEC\\NMSE (\%)$\downarrow$}
& \makecell{H\&F $\to$ POP\\NMSE (\%)$\downarrow$}
& \makecell{WTH $\to$ OPT\\MSE$\downarrow$}
& \makecell{OPT $\to$ RD\\clDice $\uparrow$} \\
\midrule
ML-4M-3B \citep{mizrahi20234m} & 77.800 & 32.300 & 106.100 & 0.063 & 0.139 \\
Show-O2-1.5B \citep{showo2} & 53.500 & \textbf{3.324 }& 164.200 & 0.299 & 0.024 \\
UniGen-1.5B \citep{tian2026unigen} & 19.000 & 96.900 & 87.9 & \textbf{0.036} & -- \\
MODUS-7B \citep{ye2026modus} & \textbf{12.397} & 37.500 & \textbf{20.229 }& 0.126 & \textbf{0.288} \\
\bottomrule
\end{tabular}%
}
\end{table*}

\FloatBarrier
\subsection{Agentic Tasks}
\vspace{-0.5em}

\subsubsection{Forecasting}
Table~\ref{tab:prediction-full600} compares agent forecasts and historical baselines across 600 cases. Three main findings follow.
(1) Agents improve carbon forecasts while consulting multimodal data. Across the five agent configurations, carbon emission MAE decreases by a median of 112.08 tonnes C/tile/year relative to the last-year baseline, a 17.334\% reduction. Tool execution records show that agents use statistical machine learning tools to interpret target history trends and, in some cases, incorporate other urban modalities and indicators (e.g., nighttime lights and satellite imagery) to refine the final estimate (Appendix case \ref{tab:carbon-case} and \ref{tab:carbon-case2}). (2) Historical averaging remains effective for LST. The ten-year mean outperforms all evaluated agents, suggesting that multi-year averaging captures persistent local thermal conditions while smoothing annual fluctuations. Agents do not consistently adapt their forecasting strategies to these temporal differences: carbon emissions may require greater emphasis on recent activity-related changes, whereas LST may benefit more from long-term historical patterns. (3) Multimodal reasoning does not consistently translate into forecast adjustments in complex tasks. According to the error analysis, we find that agents can identify relevant urban context, but often use it to check or justify a historical estimate rather than quantitatively revise their answer. The inspected cases reveal a gap between interpreting additional modalities and using them to calibrate forecasts (Appendix~\ref{app:prediction-case-study}).

\begin{table*}[!h]
\vspace{-1em}
\caption{Forecasting performance on 600 cases: 220 LST, 220 carbon,
and 160 property cases}
\label{tab:prediction-full600}
\centering
\small
\setlength{\tabcolsep}{6pt}
\renewcommand{\arraystretch}{1.08}
\resizebox{\textwidth}{!}{%
\begin{tabular}{lrrr}
\toprule
Method & LST MAE ($^\circ$C) $\downarrow$
& Carbon MAE (tonnes C/tile/year) $\downarrow$
& Property MAPE (\%) $\downarrow$ \\
\midrule
Earth Agent (GPT Luna)     & 1.436 & 548.58          & 9.47 \\
Earth Agent (DeepSeek)     & 1.400 & 511.62 & 8.64 \\
Codex (DeepSeek)  & 1.350  & \textbf{485.78} & \textbf{8.21} \\
Codex (GPT-5.5)        & 1.341 & 542.21          & 8.47 \\
OpenCode (GPT Luna)        & 1.354 & 534.51          & 9.43 \\
OpenCode (Gemini Flash Lite) & 1.470 & 584.56        & 9.59 \\
OpenCode (GLM) & 1.361  & 552.64  & 9.00    \% \\

\midrule
Last-year value ($t-1$)    & 1.677 & 646.59          & 8.67 \\
Ten-year mean             & \textbf{1.313} & 2,169.03 & 22.41 \\
\bottomrule
\end{tabular}%
}
\vspace{-1em}
\end{table*}


\begin{figure*}[t]
    \centering
    \includegraphics[width=\textwidth]{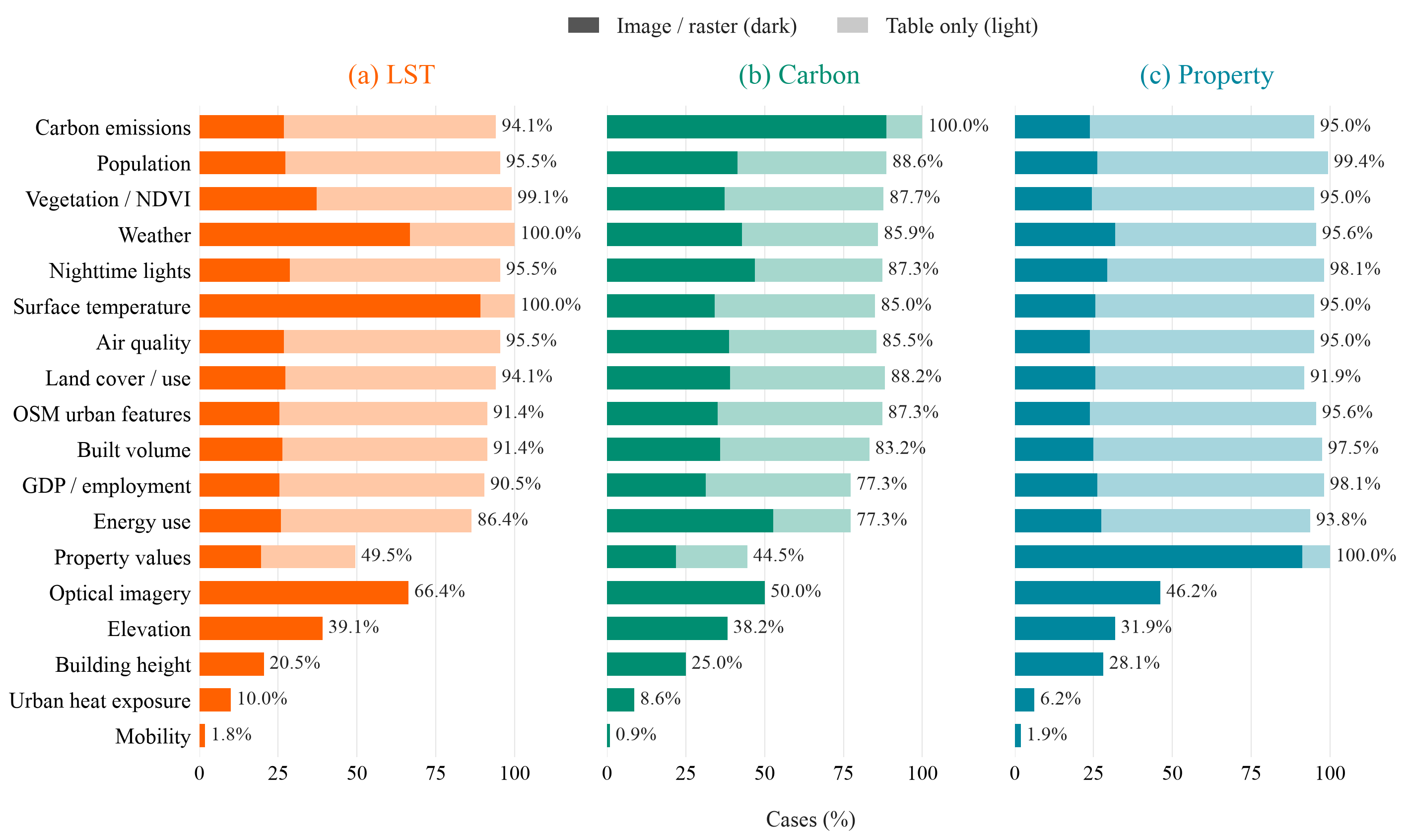}
    \caption{Modality access by Codex (DeepSeek) across forecasting tasks.
    Dark segments indicate image/raster access; light segments indicate
    table-only access. Labels show the total percentage of cases
    accessing each modality.}
    \label{fig:modality-usage}
\end{figure*}

Figure~\ref{fig:modality-usage} illustrates modality usage by Codex (DeepSeek) across three forecasting tasks. For LST, frequent access to weather data and optical imagery is consistent with the influence of climatic conditions and surface characteristics on temperature, while vegetation and built form provide complementary information about cooling and heat retention, respectively. For carbon emissions, the agent accesses energy-use rasters more frequently, indicating attention to emission-related activity. Nighttime lights and population provide indirect proxies whose relationship with emissions depends on energy intensity and the composition of emission sources. For property prices, the agent accesses population, economic indicators, and built volume, covering potential demand and supply signals. Overall, the figure shows that the agent selects different evidence according to the task. Future agents could better exploit multimodal information and improve forecasting by moving beyond identifying relevant modalities toward reasoning about their complementary, overlapping, or offsetting effects and translating these relationships into quantitative adjustments.

\subsubsection{Planning}
\vspace{-0.5em}
\begin{table*}[t]
\centering
\small
\caption{Agentic planning results for 150 EV charging port cases and 100 school capacity cases. The school balanced rate uses a 95--105\% enrollment-to-capacity range. }
\label{tab:agentic_planning}
\setlength{\tabcolsep}{6pt}
\begin{tabular}{lcccc}
\toprule
& \multicolumn{2}{c}{EV charging}
& \multicolumn{2}{c}{School capacity} \\
\cmidrule(lr){2-3}\cmidrule(lr){4-5}
Method & Accuracy (\%) & Macro-F1
& Balance rate (\%) & Mean balance \\
\midrule
Constant-class baseline  & 25.3 & 0.101 & -- & -- \\
Three-year-mean baseline & -- & -- & 84.0 & 96.60 \\
\midrule
Earth Agent (GPT Terra) & 35.3 & 0.339 & 81.0 & 96.91 \\
Earth Agent (DeepSeek)  & \textbf{42.0} & 0.382 & 88.0 & 97.50 \\
Codex (DeepSeek)        & 38.0 & 0.366
                       & \textbf{89.0} & \textbf{97.61} \\
Codex (GPT-5.5)         & 40.7 & \textbf{0.387} & 84.0 & 97.20 \\
OpenCode (GPT Terra)    & 36.0 & 0.333 & 80.0 & 96.74 \\
OpenCode (Gemini)       & 33.3 & 0.282 & 84.0 & 97.45 \\
\bottomrule
\end{tabular}
\end{table*}
Table~\ref{tab:agentic_planning} summarizes both planning tasks. For EV charger expansion planning, agents outperform the constant-class baseline by up to 16.7 percentage points in accuracy. Successful cases combine multimodal data such as population and POI with charging histories to support decisions (Table~\ref{tab:agent-appendix-ev-confusion}). However, agents often underestimate the real requirement due to focusing too much on the historical trend instead of considering the increasing need for EV facilities as cities develop. For school capacity planning, agents improve balance rate over the three-year average baseline by up to 5 percentage points in balance rate, showing the agent's ability to translate recent enrollment trends into capacity adjustments. Incorporating more domain-specific knowledge (e.g., age structure) has the potential to improve agents' task-specific ability \citep{dfe_school_capacity_202425}.

\subsubsection{Resource Analysis}
Table~\ref{tab:food_access} reports performance on 150 food-access analysis cases, each containing four spatial cells' prediction. Codex (DeepSeek) achieves the highest macro-F1, but its exact-case accuracy remains low, mainly due to difficulty in correctly resolving all four cells within a case. Identifying joint low-income and low-access areas also remains challenging for some configurations. For example, Gemini correctly identifies only 2 of 62 such cells. Inspected agent rationales show that commercial activity is sometimes treated as evidence of supermarket accessibility, overlooking differences between general urban activity and access to specific services. These cases highlight limitations in interpreting modality semantics and spatial scale. More effective multimodal reasoning requires distinguishing direct evidence from indirect proxies and checking whether their spatial coverage supports the requested judgments.

\begin{table}[!htbp]
\centering
\small
\caption{Food access classification on 150 cases.
Exact-case accuracy requires all four cells on one tile to be correct.}
\label{tab:food_access}
\setlength{\tabcolsep}{5pt}
\begin{tabular}{lccc}
\toprule
Method & Cell acc. (\%) & Macro-F1 & Exact-case acc. (\%) \\
\midrule
Majority-class baseline & 49.67 & 0.221 & 32.0 \\
\midrule
Earth Agent (GPT Terra) & 50.00 & 0.385 & 33.3 \\
Earth Agent (DeepSeek)  & 51.33 & 0.385 & 14.7 \\
Codex (DeepSeek)        & 55.33 & \textbf{0.498} & 18.7 \\
Codex (GPT-5.5)         & 53.83 & 0.475 & 37.3 \\
OpenCode (GPT Terra)    & 53.67 & 0.429 & 35.3 \\
OpenCode (Gemini)       & \textbf{57.00} & 0.412 & \textbf{38.7} \\
\bottomrule
\end{tabular}
\end{table}
\vspace{-0.5em}

Tool-use traces show that Earth Agent relies more on specialized statistical and machine-learning tools for forecasting, while Codex and OpenCode primarily analyze data through code (Appendix~\ref{app:agent-tool-use}). More tool calls do not consistently correspond to better performance. For example, in food-access analysis, Earth Agent (DeepSeek) averages 28.520 calls per case, compared with 12.907 for Codex (GPT-5.5), yet achieves lower macro-F1 (0.385 versus 0.475) and exact-case accuracy (14.7\% versus 37.3\%). The inspected cases illustrate how historical window selection, capacity margins, and spatial interpretation shape the final decisions. These observations highlight the importance of how agents interpret and apply retrieved evidence, beyond the volume of tool use.
\FloatBarrier
\section{Conclusion}
\vspace{-0.5em}

We present  AgenticCity-Bench, a dataset that integrates 18 urban modalities across 16 cities from 2000 to 2025. It provides a foundation for research on urban understanding and agentic analysis. Our experiments demonstrate the potential for cross-modal translation of urban data, while highlighting the need to better balance different modalities. We further evaluate six agentic city tasks across three categories with 1,000 cases and identify both promising capabilities and persistent limitations. The gap between understanding and action, insufficient domain knowledge, and improper tool use can be improved in the future. Researchers can build on our dataset and benchmark to develop more effective urban foundation models and urban agents to provide suggestions to policymakers and advance the development of urban science and the Earth domain.



\clearpage

\section{AI Disclosure}

The authors declare that artificial intelligence (AI) tools, specifically ChatGPT, were used solely to assist with language polishing and grammar checking of the manuscript text. All intellectual content, data analysis, interpretations, and conclusions were conceived, written, and verified by the authors. We take responsibility for the final content of this work, including text, claims or artifacts produced with the aid of generative AI.

\section{Acknowledgements}
This research is supported by the National Research Foundation (NRF), Prime Minister’s Office, Singapore under its Campus for Research Excellence and Technological Enterprise (CREATE) programme. The Mens, Manus, and Machina (M3S) is an interdisciplinary research group (IRG) of the Singapore MIT Alliance for Research and Technology (SMART) centre. 



\begingroup
\small
\bibliography{refs}
\bibliographystyle{iclr2027_conference}
\endgroup

\clearpage
\appendix
\makeatletter
\setlength{\@fptop}{0pt}
\makeatother

\section{Related Work}
\label{app:related_work}

\paragraph{Multimodal and Temporal Urban and EO Benchmarks}
Research on EO and urban benchmarks can be organized around three complementary goals: learning from annotated scenes, transferring across heterogeneous observations, and understanding urban conditions across modalities and time
\citep{xia2018dota,nedungadi2024mmearth,xuan2025dynamicvl,
feng2025urbanllava}.
Early task-focused resources established reusable supervision: DOTA provided oriented object annotations for aerial detection, while RSVQA paired remotely sensed imagery with natural-language questions and answers \citep{xia2018dota,lobry2020rsvqa}.
Subsequent resources broadened both training and evaluation, with geographically aligned multimodal pretraining data and curated task suites for assessing transferable representations \citep{nedungadi2024mmearth,lacoste2023geobench}.
Temporal benchmarks introduced a different axis of coverage, progressing
from monthly building tracking to multi-year analyses of urban transformation \citep{vanetten2021spacenet7,xuan2025dynamicvl}. In parallel, urban data pipelines and instruction benchmarks have connected imagery with structured geospatial and mobility data, extending evaluation beyond surface appearance to cross-modal urban understanding \citep{wang2024urbandatalayer,feng2025urbanllava}.

These advances highlight the challenge of jointly modeling heterogeneous urban data across modalities and time \citep{wang2024urbandatalayer,feng2025urbanllava,xuan2025dynamicvl}. Generative EO models can synthesize missing modalities from available observations, but integrating environmental observations with urban activity records over time remains challenging \citep{jakubik2025terramind}.

\vspace{-1em}
\paragraph{Geospatial Agents for Urban Analysis and Planning} Agentic geospatial research brings together simulation-based decision environments, language-guided analytical workflows, and increasingly process-aware evaluation
\citep{zhang2019cityflow,akinboyewa2025giscopilot,
acbEarthAgent2026,lai2026ustbench}.
Before the rise of LLM agents, CityFlow already provided a large-scale traffic simulation environment and interfaces for multi-agent
reinforcement learning, establishing an action-and-feedback setting for urban control \citep{zhang2019cityflow}. In the broader language-agent literature, ReAct subsequently articulated how interleaving reasoning with actions allows models to obtain external information and update their plans  \citep{yao2023react}. Domain-specific agentic systems have adapted these capabilities to GIS workflow
generation, multimodal scientific analysis, and the coordination of retrieval, perception, and reasoning tools
\citep{akinboyewa2025giscopilot,acbEarthAgent2026,xiao2026geomm}. Urban evaluation has developed alongside this analytical strand: CityBench combines heterogeneous city data with interactive simulation, while USTBench diagnoses understanding, forecasting, planning, and
reflection within urban tasks \citep{feng2025citybench,lai2026ustbench}.

These developments make it possible to examine how agents reach their results \citep{acbEarthAgent2026,lai2026ustbench}. However, failures in tool use, domain-specific reasoning, and long-horizon adaptation show that successful perception does not guarantee reliable analysis or decision-making \citep{acbEarthAgent2026,xiao2026geomm,lai2026ustbench}. Agents must therefore select relevant information, identify missing prerequisites, and construct spatial plans under explicit constraints.

\section{Benchmark Illustration}
\label{app:benchmark}
\subsection{Study Regions and Grid Geometry}
\label{app:cities}
The benchmark covers 16 city-centered regions selected to span six inhabited continents and different urban forms, climates, and data environments. Each
region is projected to a local metric coordinate reference system before
tiling. Separate, non-overlapping grids are constructed at 2-km and 5-km scales; the 5-km grid is not produced by aggregating the 2-km grid. A tile is identified by city, scale, row, and column, which provides a stable spatial key for joining heterogeneous sources without implying that their native resolutions are identical. Table~\ref{tab:cities} reports the projected grid
geometry and the number of tiles retained at each scale.

\begin{table}[!htbp]
\caption{Multi-scale city-grid inventory across 16 cities. Grid dimensions are
rows$\times$columns and parenthetical values give cell counts. Each scale uses
its own projected grid and is evaluated separately.}
\label{tab:cities}
\centering
\small
\setlength{\tabcolsep}{4pt}
\renewcommand{\arraystretch}{1.04}
\begin{tabular*}{\textwidth}{@{\extracolsep{\fill}}llccc@{}}
\toprule
City & Continent & CRS & \twokm grid (cells) & \fivekm grid (cells) \\
\midrule
Amsterdam & Europe & EPSG:32631 & $10\times14$ (140) & $3\times4$ (12) \\
Boston & North America & EPSG:32619 & $13\times13$ (169) & $4\times4$ (16) \\
Busan & Asia & EPSG:32652 & $16\times13$ (208) & $5\times5$ (25) \\
Cape Town & Africa & EPSG:32734 & $14\times13$ (182) & $4\times5$ (20) \\
Christchurch & Oceania & EPSG:32759 & $30\times32$ (960) & $11\times12$ (132) \\
Washington, D.C. & North America & EPSG:32618 & $13\times11$ (143) & $5\times4$ (20) \\
London & Europe & EPSG:32630 & $24\times30$ (720) & $9\times11$ (99) \\
Lyon & Europe & EPSG:32631 & $11\times13$ (143) & $3\times4$ (12) \\
New York City & North America & EPSG:32618 & $26\times24$ (624) & $10\times9$ (90) \\
San Francisco & North America & EPSG:32610 & $17\times19$ (323) & $6\times6$ (36) \\
Santiago & South America & EPSG:32719 & $13\times14$ (182) & $4\times5$ (20) \\
Shanghai & Asia & EPSG:32651 & $13\times13$ (169) & $5\times4$ (20) \\
Singapore & Asia & EPSG:32648 & $20\times32$ (640) & $7\times12$ (84) \\
Sydney & Oceania & EPSG:32756 & $13\times13$ (169) & $5\times4$ (20) \\
Tokyo & Asia & EPSG:32654 & $17\times17$ (289) & $6\times6$ (36) \\
XiongAn & Asia & EPSG:32650 & $25\times31$ (775) & $9\times12$ (108) \\
\midrule
Total & Six continents & - & 5,836 &  750 \\
\bottomrule
\end{tabular*}
\end{table}

\subsection{Modality Sources}
\label{app:source}

The 18 semantic modalities combine satellite observations, derived EO
products, geographic databases, model-based estimates, and city-specific official records. Table~\ref{tab:source} records the primary data source and access for each modality.

\begin{table*}[p]
\caption{Primary source registry for the AgenticCity-Bench modalities.}
\label{tab:source}
\centering

\scriptsize
\Urlmuskip=0mu plus 1mu\relax
\setlength{\tabcolsep}{3pt}
\renewcommand{\arraystretch}{1.10}
\begin{tabularx}{\textwidth}{@{}
>{\raggedright\arraybackslash}p{0.8in}
>{\raggedright\arraybackslash}p{1.00in}
>{\raggedright\arraybackslash}X@{}}
\toprule
Modality & Dataset name & Source website \\
\midrule
\texttt{Optical Imagery} & Landsat Collection 2 Level-2& \url{https://www.usgs.gov/landsat-missions/landsat-collection-2-level-2-science-products} \\

\texttt{Optical Imagery} & Planet RapidEye(2009--2015) and Mosaics (2016--2025) & \url{https://docs.planet.com/data/imagery/basemaps/} \\
\texttt{Roads} & Open Street Map& \url{https://planet.openstreetmap.org/planet/full-history/} \\
\texttt{POI} & Open Street Map& \url{https://www.openstreetmap.org/} \\

\texttt{GPS} & Open Street Map& \url{https://api.openstreetmap.org/api/0.6/trackpoints} \\
\texttt{Height \& Footprint} & 3D-GloBFP global building-footprint height & \url{https://essd.copernicus.org/articles/16/5357/2024/} \\

\texttt{DEM} & Copernicus DEM GLO-30 & \url{https://dataspace.copernicus.eu/explore-data/data-collections/copernicus-contributing-missions/collections-description/COP-DEM} \\

\texttt{Built Volume} & GHSL GHS-BUILT-V R2023A & \url{https://data.jrc.ec.europa.eu/dataset/ab2f107a-03cd-47a3-85e5-139d8ec63283} \\

\texttt{Population} & GHSL GHS-POP R2023A & \url{https://data.europa.eu/data/datasets/2ff68a52-5b5b-4a22-8f40-c41da8332cfe?locale=en} \\

\texttt{NDVI} & Landsat Collection 2 Level-2& \url{https://www.usgs.gov/landsat-missions/landsat-collection-2-level-2-science-products} \\

\texttt{Heat} & Landsat Collection 2 Level-2 Surface Temperature & \url{https://www.usgs.gov/landsat-missions/landsat-collection-2-surface-temperature} \\

\texttt{Land Cover} & GLC\_FCS30D & \url{https://zenodo.org/records/8239305} \\

\texttt{Land Use} & HILDA+ v2& \url{https://doi.org/10.1594/PANGAEA.974335} \\

\texttt{NO$_2$} & Surface-no2 & \url{https://sites.wustl.edu/acag/datasets/surface-no2/} \\

\texttt{PM$_{2.5}$} & SatPM2.5 & \url{https://sites.wustl.edu/acag/surface-pm2-5/} \\

\texttt{Weather} & CHELSA v2.1 & \url{https://www.chelsa-climate.org/datasets} \\

\texttt{Weather} & Daymet & \url{https://daymet.ornl.gov/getdata} \\

\texttt{Carbon Emissions} & ODIAC& \url{https://www.odiac.org/data-product.html} \\

\texttt{Nighttime Lights} & VIIRS nighttime lights (VNL) & \url{https://eogdata.mines.edu/products/vnl/} \\

\texttt{Property} & City-specific property valuation records, harmonized to 2017 PPP international dollars & \url{https://api.worldbank.org/v2/country/all/indicator/FP.CPI.TOTL?format=json}\newline \url{https://api.worldbank.org/v2/country/all/indicator/PA.NUS.PPP?format=json}\newline \url{https://gisportal.boston.gov/arcgis/rest/services/Assessing}\newline \url{https://www.data.go.kr/data/15126469/openapi.do}\newline \url{https://www.capetown.gov.za/}\newline \url{https://services.arcgis.com/neT9SoYxizqTHZPH/arcgis/rest/services/OCFO_ITSPE_view_05212026/FeatureServer/53}\newline \url{https://www.gov.uk/government/statistical-data-sets/price-paid-data-downloads}\newline \url{https://files.data.gouv.fr/geo-dvf/latest/}\newline \url{https://data.cityofnewyork.us/Housing-Development/Annualized-Rolling-Sales-Update/w2pb-icbu}\newline \url{https://data.sfgov.org/Housing-and-Buildings/Assessor-Historical-Secured-Property-Tax-Rolls/wv5m-vpq2}\newline \url{https://data.gov.sg/collections/189}\newline \url{https://www.ura.gov.sg/property-data/private-residential-properties}\newline \url{https://www.nsw.gov.au/housing-and-construction/land-values-nsw/resource-library/property-sales-data-guide}\newline \url{https://www.reinfolib.mlit.go.jp/} \\

\texttt{Electricity} & City-specific official records. & \url{https://data.boston.gov/dataset/building-emissions-reduction-and-disclosure-ordinance}\newline \url{https://citymaps.capetown.gov.za/agsext/rest/services/Theme_Based/ODP_SPLIT_10/MapServer/12}\newline \url{https://www.emi.ea.govt.nz/Retail/Reports/0YUCE0}\newline \url{https://www.gov.uk/government/collections/sub-national-electricity-consumption-data}\newline \url{https://data.london.gov.uk/dataset/london-energy-and-greenhouse-gas-inventory-leggi-2ko63}\newline \url{https://www.data.gouv.fr/datasets/donnees-locales-de-consommation-delectricite-de-gaz-naturel-et-de-chaleur-et-de-froid-iris-a-partir-de-2018}\newline \url{https://data.cityofnewyork.us/Environment/NYC-Building-Energy-and-Water-Data-Disclosure-for-/5zyy-y8am}\newline \url{https://data.sfgov.org/Energy-and-Environment/Existing-Buildings-Benchmark-Reports/4ua7-5sfx/about_data}\newline \url{https://www.ine.gob.cl/herramientas/portal-de-mapas/siedu}\newline \url{https://www.ema.gov.sg/resources/singapore-energy-statistics/chapter3}\newline \url{https://data.cityofsydney.nsw.gov.au/datasets/electricity-consumption-by-suburb}\newline \url{https://www.kankyo.metro.tokyo.lg.jp/documents/d/kankyo/co2-reduction-dashboard_source_202508}\newline \\
\bottomrule
\end{tabularx}
\end{table*}

\subsection{ Data Availability}
\label{app:data_availability}
\begin{figure}
    \centering
    \includegraphics[width=1\linewidth]{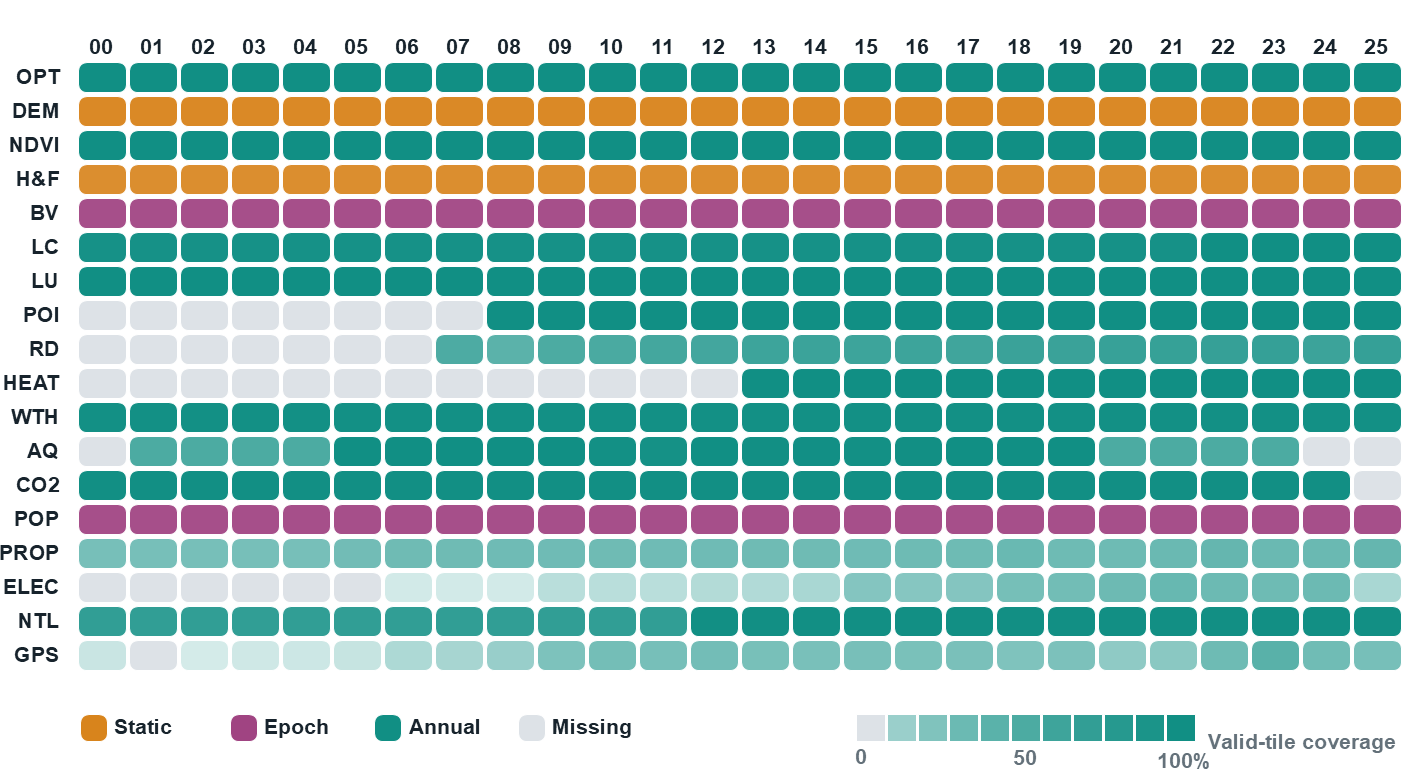}
    \caption{\textbf{Temporal availability of the 18 benchmark modalities from 2000 to 2025.}
Columns denote calendar years and rows denote modalities. Color intensity indicates the proportion of valid tiles across the 16 cities. }
    \label{fig:temporal_coverage}
\end{figure}

Coverage is summarized at two distinct levels. Figure~\ref{fig:temporal_coverage} and Tables~\ref{tab:modality_availability},\ref{tab:city_availability} describe coverage for each city, year, and modality. A combination is considered available when the release contains at least one valid tile for the specified city, calendar year, and semantic modality. 

Across the 16 cities and 26 years from 2000 to 2025, the dataset includes 18 modalities for the any-to-any translation tasks, indexed by city, year, and modality. In addition, a grid-context modality is provided only during training to supply spatial background information and is not used as a source or target in the translation tasks. The grid context is available for all cities and is therefore excluded from the modality coverage analysis. Static and epoch-based data are reused across the corresponding years when applicable, rather than being treated as independent annual observations.
For the 18 translation modalities, there are $16 \times 26 \times 18 = 7{,}488$ possible city, year, and modality combinations in Table \ref{tab:modality_availability} . Among these, 6,309 combinations are available, corresponding to an overall availability rate of 84.3\%. The main gaps arise from electricity, property, air quality, GPS, POI, and historical road-network records in Fig \ref{fig:temporal_coverage}.

\begin{table}[!t]
\caption{Spatial granularity and temporal coverage of the 18 semantic modalities in AgenticCity-Bench.}
\label{tab:modality-spatiotemporal}
\centering
\scriptsize
\setlength{\tabcolsep}{5pt}
\begin{tabularx}{\linewidth}{L{1.65in}L{1.65in}X}
\toprule
Semantic modality & Spatial granularity & Temporal coverage \\
\midrule

Built volume &
$\sim$100 m &
2000, 2005, 2010, 2015, 2020, 2025 (5-year epochs) \\

Road network &
Vector geometry / rasterized grid &
Source-specific historical snapshots \\

Land cover &
30 m &
2000--2025 \\

Land use &
Source raster grid &
2000--2025 \\

Building height \& footprint &
Building / raster grid &
Static \\

Points of interest &
Point / building-level raster &
Source-specific historical snapshots \\

\addlinespace
Electricity &
Building / point / administrative unit &
City-specific years \\

Nighttime lights &
$\sim$500 m &
2012--2025; source-specific historical coverage \\

Carbon emissions &
1 km &
2000--2024 (annual) \\

Mobility (GPS) &
GPS sampling points / tile-level density &
Source-specific years \\

Population &
$\sim$1 km &
2000, 2005, 2010, 2015, 2020, 2025 (5-year epochs) \\

Property value &
Parcel / point / postcode / neighborhood &
City-specific years \\

\addlinespace
Optical imagery &
$\sim$30 m (Landsat); $\sim$5 m (Planet) &
2000--2025 \\

Terrain &
$\sim$30 m &
Static \\

Vegetation &
$\sim$30 m &
2000--2025 (annual) \\

Weather &
$\sim$1 km &
2000--2025 (annual aggregation) \\

Surface temperature &
Landsat-derived spatial support &
2000--2025; source-dependent annual/composite observations \\

Air quality &
Source-specific coarse grid &
Source-specific annual coverage \\

\bottomrule
\end{tabularx}

\vspace{2pt}
\footnotesize
\textit{Note.}Epoch-based and static
data may be reused across calendar years, but such reuse does not constitute independent annual observations.
\end{table}

\begin{table}[t]
\centering
\small
\caption{Data availability by modality in the current 2-km release.
Each modality has $16 \times 26 = 416$ possible city-year combinations.
Static and epoch-based data are counted as available when explicitly
reused as model inputs.}
\label{tab:modality_availability}
\begin{tabular}{lrr}
\toprule
\textbf{Modality} & \textbf{Available combinations.} & \textbf{Availability} \\
\midrule
Optical imagery       & 416 / 416 & 100.000\% \\
Terrain               & 416 / 416 & 100.000\% \\
Vegetation            & 416 / 416 & 100.000\% \\
Footprint \& height  & 364 / 416 & 87.500\%  \\
Built volume          & 416 / 416 & 100.000\% \\
Land cover            & 416 / 416 & 100.000\% \\
Land use              & 416 / 416 & 100.000\% \\
POI                   & 288 / 416 & 69.231\%  \\
Road network          & 300 / 416 & 72.115\%  \\
Heat         & 358 / 416 & 86.058\%  \\
Weather               & 416 / 416 & 100.000\% \\
Air quality           & 240 / 416 & 57.692\%  \\
Carbon emissions      & 400 / 416 & 96.154\%  \\
Population            & 416 / 416 & 100.000\% \\
Property economy      & 232 / 416 & 55.769\%  \\
Electricity           & 111 / 416 & 26.683\%  \\
Night-time lights     & 404 / 416 & 97.115\%  \\
GPS density           & 284 / 416 & 68.269\%  \\
\midrule
\textbf{Overall}
& \textbf{6,309 / 7,488}
& \textbf{84.255\%} \\
\bottomrule
\end{tabular}
\end{table}

\begin{table}[t]
\centering
\small
\caption{Data availability by city.
Each city has $26 \times 18 = 468$ possible year-modality combinations.
}
\label{tab:city_availability}
\begin{tabular}{lr@{\hspace{1.2cm}}lr}
\toprule
\textbf{City} & \textbf{Availability}
& \textbf{City} & \textbf{Availability} \\
\midrule
Amsterdam
& 392 / 468 (83.761\%)
& Boston
& 413 / 468 (88.248\%) \\

Busan
& 403 / 468 (86.111\%)
& Cape Town
& 395 / 468 (84.402\%) \\

Christchurch
& 353 / 468 (75.427\%)
& Washington, D.C.
& 415 / 468 (88.675\%) \\

London
& 420 / 468 (89.744\%)
& Lyon
& 390 / 468 (83.333\%) \\

New York City
& 411 / 468 (87.821\%)
& San Francisco
& 411 / 468 (87.821\%) \\

Santiago
& 379 / 468 (80.983\%)
& Shanghai
& 316 / 468 (67.521\%) \\

Singapore
& 418 / 468 (89.316\%)
& Sydney
& 420 / 468 (89.744\%) \\

Tokyo
& 401 / 468 (85.684\%)
& Xiong'an
& 372 / 468 (79.487\%) \\
\midrule
\multicolumn{2}{l}{\textbf{Overall}}
& \multicolumn{2}{r}{\textbf{6,309 / 7,488 (84.255\%)}} \\
\bottomrule
\end{tabular}
\end{table}

\subsection{Preprocessing and Alignment}
\label{app:processing_alignment}

Data availability is also determined using finite-value checks, data-specific no-data values, and coverage masks when available. Image modalities require at least 98\% valid pixels, while numerical modalities require valid source support above a predefined threshold. If a modality does not meet these criteria, it is marked as unavailable for the corresponding city, year, tile combination.

Missing observations are represented using explicit availability masks and are neither zero-filled nor replaced with synthetic observations. For example, the electricity data for Washington DC do not cover the surrounding areas and suburbs. Tiles outside the reported coverage are therefore marked as unavailable rather than assigned zero consumption. Conversely, zero is retained when it represents a meaningful observation, including no road in \texttt{road network}, ocean in Land Use, and physical zeros in continuous fields.

Tables~\ref{tab:visual-modalities} and~\ref{tab:statistical-modalities} summarize the fields used for the 18 urban modalities, their processing, and the corresponding units.

\begin{table*}[t]
\caption{Visual spatial modalities used by any2any models. Each source GeoTIFF is converted to a tile-aligned PNG with a modality-specific fixed rendering rule.}
\label{tab:visual-modalities}
\centering
\scriptsize
\setlength{\tabcolsep}{3pt}
\renewcommand{\arraystretch}{1.12}
\begin{tabularx}{\textwidth}{@{}
>{\raggedright\arraybackslash}p{1.25in}
>{\raggedright\arraybackslash}p{1.62in}
>{\raggedright\arraybackslash}X
>{\raggedright\arraybackslash}p{0.78in}@{}}
\toprule
Modality & Source field(s) & MODUS representation and processing & Unit \\
\midrule
Landsat optical imagery &
B1=red, B2=green, B3=blue &
Native RGB &
     \\

Planet optical imagery &
B1=red, B2=green, B3=blue &
Native RGB & - \\

Road network &
B1=major-road raster &
Fixed white-background/red-road image &
- \\

Height\& footprint & 
B1=\texttt{height\_m} &
Fixed-range grayscale PNG: black is zero/background; brighter gray indicates taller buildings. Zero is a valid height. &
m \\

DEM &
B1=\texttt{elevation\_m} &
Train-global linear grayscale \texttt{dem.png}. Zero is valid elevation. & m \\

NDVI &
B1=\texttt{NDVI} &
Fixed-range grayscale PNG: darker values indicate lower NDVI and brighter values indicate higher NDVI &
1 \\

Land cover &
B1=raw land-cover class ID &
Fixed categorical-palette PNG of B1 land-cover classes &
Class ID \\

Land use &
B1=raw HILDA class ID &
Fixed categorical-palette PNG of B1 land use classes; class 99 is no-data. &
Class ID \\

POI  &
B1= class ID &
Fixed categorical-palette PNG of B1 POI classes. &
Class ID \\

Built volume &
B1= total built-volume density &
\texttt{log1p}-scaled grayscale PNG of total built-volume density. &
m$^3$/m$^2$ \\
\bottomrule
\end{tabularx}
\vspace{2pt}
\parbox{\textwidth}{\scriptsize
\textit{Note.} B1, B2, and B3 denote the first, second, and third bands of the source GeoTIFF, respectively; band meanings are modality-specific.}
\end{table*}

\begin{table*}[t]
\caption{Statistical and geo-temporal text modalities used by MODUS. Numerical
fields are serialized as typed JSON within the prompt or target text. All
raster summaries are computed over valid source support only.}
\label{tab:statistical-modalities}
\centering
\scriptsize
\setlength{\tabcolsep}{3pt}
\renewcommand{\arraystretch}{1.12}
\begin{tabularx}{\textwidth}{@{}
>{\raggedright\arraybackslash}p{1.35in}
>{\raggedright\arraybackslash}p{1.78in}
>{\raggedright\arraybackslash}X
>{\raggedright\arraybackslash}p{0.95in}@{}}
\toprule
Modality & JSON field(s) & Tile-level processing & Unit \\
\midrule
Population\_GHSL &
\texttt{population\_density} &
Area-weighted mean of the population-density display field. &
persons/km$^2$ \\

Air quality NO2 &
\texttt{annual\_mean\_no2} &
Area-weighted mean of annual surface NO$_2$. &
ppb \\

Air quality PM25 &
\texttt{annual\_mean\_pm25} &
Area-weighted mean of annual PM$_{2.5}$. &
$\mu$g/m$^3$ \\

Temperature\_LandsatST &
\texttt{mean\_celsius}; \texttt{median\_celsius} &
Area-weighted means of source B1 (mean) and B2 (median), converted to Celsius. Source p90, p95, and standard-deviation bands are excluded in V1. &
$^\circ$C \\

Weather &
\texttt{tmin\_c}; \texttt{tmax\_c}; \texttt{tmean\_c}; \texttt{prcp\_mm}; \texttt{srad\_w\_m2}; \texttt{rh\_fraction} &
The six shared fields are area-weighted over valid source cells. &
\texttt{tmin\_c}, \texttt{tmax\_c}, \texttt{tmean\_c}: $^\circ$C; \texttt{prcp\_mm}: mm; \texttt{srad\_w\_m2}: W/m$^2$; \texttt{rh\_fraction}: \% \\

Property value &
\texttt{property\_mean\_2017ppp} &
Area-weighted mean property value, expressed in 2017 PPP international dollars  &
2017 PPP international dollars per property observation \\

Electricity &
\texttt{electricity} &
Valid source-cell electricity values are summed and divided by tile area. &
kWh/km$^2$ \\

CO$_2$ &
\texttt{carbon} &
Valid source-cell carbon emissions are summed and divided by tile area. &
tonne C/km$^2$/year \\

Nighttime lights &
\texttt{radiance} &
Area-weighted mean annual nighttime radiance. &
nW/cm$^2$/sr \\

OSM\_GPS\_Trace &
\texttt{gps\_trace\_density\_log1p};  &
Aggregated to the tile from GPS-trace point records. The primary value is $\log(1+\mathrm{points/km^2})$. &
log1p(points/km$^2$) \\

grid &
City, country, year, scale, row, column, and WGS84 bounds &
Automatic geo-temporal text context. It is always included in the prompt and never selected as a generation target. &
-- \\
\bottomrule
\end{tabularx}
\end{table*}

\subsubsection{Function modalities Preprocessing}
\label{app:function_modality}

\paragraph{Road network (OSM).}
In Table \ref{tab:visual-modalities}, we use band 1 of the supplied OSM raster as the road layer. Nonzero pixels are rendered in a fixed red RGB value $(224,64,69)$ on a white background. Zero pixels indicate no mapped road and remain valid. Therefore, a tile without mapped roads can still meet the 98\% validity threshold. The release converter does not add road buffers, dilate road lines, or create new road pixels. Each raster keeps the year or snapshot provided with the original OSM data. A blank road image indicates that no roads are mapped in the supplied data, not that the area has no roads in reality.

\paragraph{Building height \& footprint (H\&F).}
In Table \ref{tab:visual-modalities}, band 1 of BuildingHeight\_3DGloBFP provides building height in metres. We treat it as static built-environment information, even when the same data appear in multiple year directories. Valid zero values are retained, and an image is included only when at least 98\% of its pixels are valid. Heights are mapped linearly to a fixed grayscale range and replicated across the three RGB channels, then saved as a PNG image. The raster encodes height values, including zero-height pixels, rather than providing a separate binary building-footprint layer. The spatial pattern of nonzero pixels may therefore show where buildings are present, but these pixels are not extracted or converted into a separate footprint layer. The PNG preserves the relative height values within the display range, but may lose the precision and extreme values of the original floating-point data.

\paragraph{Built volume (BV).}
Built volume (GHSL) is provided as 5-year epoch data. We use band 1,  which in Table \ref{tab:visual-modalities} represents total built-volume density in m$^3$/m$^2$, and exclude the non-residential band. Zero values are valid and indicate no built volume, while the explicit $-9999$ value is treated as invalid. After applying the 98\% validity filter, we apply a log transform and map the values to a fixed percentile-based grayscale range to produce a PNG image. We use the spatial grid provided in the original GHSL data without additional spatial reconstruction or resampling. According to previous works \citep{tesselaar2023flood,yu2022forest}, each 5-year epoch value is reused for the years covered by that epoch and does not represent separate yearly observations.

\paragraph{Land use (LU).}
In Table \ref{tab:visual-modalities}, band 1 of LU (HILDA) provides categorical land-use IDs. The fixed legend includes ocean, urban land, crop systems, pasture, forest types, unmanaged grass or shrubland, sparse vegetation, and inland water. ID 0 is valid and represents ocean, while ID 99 represents nodata. IDs outside the defined classes are treated as invalid. Images are included only when at least 98\% of the pixels are valid and are saved as a PNG image. The class IDs represent discrete categories rather than continuous values, so we use a fixed RGB lookup table instead of continuous interpolation. During generation, each predicted pixel is mapped to the class with the nearest predefined class color.

\paragraph{Building-function POI.}
POI refers to OSM-derived points of interest and building-function categories rasterized to the study grid. In Table \ref{tab:visual-modalities}, band 1 of POI is a 13-class building-function raster. ID 0 denotes outside/no building, while ID 1 denotes an unlabeled building. IDs 2--12 represent residential, commercial/retail/office, education, healthcare, food service, civic/public, transport/parking, tourism/leisure, industrial/warehouse, religious, and other functions. Both 0 and 1 are valid classes, not missing data. The converter uses the supplied rasterized footprint/function map, applies the 98\% valid-pixel threshold, and writes a PNG image using a fixed class palette. It does not reclip polygons or infer labels for unlabeled buildings. Temporal meaning follows the supplied OSM snapshot, and the absence of a class is distinct from missing map data. During generation, each predicted pixel is mapped to the class with the nearest predefined class color.

\paragraph{Land cover and green-space composition (LC).}
In Table \ref{tab:visual-modalities}, land cover uses the original band-1 IDs. A fixed legend retains crop, forest, shrub, grass,
wetland, impervious, bare, water, and snow/ice distinctions and maps them
to a PNG image. Thus the image also supports analysis of
green-space composition. The current legend accepts ID 0 as
non-green or unclassified; this must not be interpreted as
the road-background class or automatically as verified non-green land.
Accepted images require 98\% valid pixels. Temporal assignments are inherited; class IDs are not continuously normalized or interpolated. During generation, each predicted pixel is mapped to the class with the nearest predefined class color.

\subsubsection{Activity modalities Preprocessing}

\paragraph{Population (POP).}
Population (GHSL) provides population density in persons/km$^2$.  In Table \ref{tab:statistical-modalities}, it provides population density rather than population counts for individual pixels. We use the spatial mean of valid pixels within each tile to obtain the tile-level population density. At least 50\% of the cells must be valid, and zero population is retained as a valid value. Population data are provided for specific source epochs, and the same value is reused for the corresponding years in the dataset without additional temporal interpolation \citep{yu2022forest,tesselaar2023flood}. A log transform is applied before standardization.

\paragraph{Property economy (PROP).}

Property values are obtained from two sources.  For most cities, 
we use the existing property-value datasets listed in 
Table~\ref{tab:source}. For Shanghai and Cape Town, where comparable gridded datasets are unavailable, records are collected from city-specific property websites.

For existing raster data, we use the mean value. A pixel is valid when it is finite and is not excluded by the source nodata value or raster mask. Let $x_i$ be the stored value of pixel $i$ and 
$m_i\in\{0,1\}$ its validity indicator. The valid-support ratio is $
r_{\mathrm{valid}}=\frac{\sum_i m_i}{N},$ 
where $N$ is the number of pixels in the tile. A tile is retained when 
$r_{\mathrm{valid}}\geq0.5$, and its property value is computed as $x_{\mathrm{PROP}}=\frac{\sum_i m_i x_i}{\sum_i m_i}. $

 All monetary values are converted from the reported local currency and year to 2017 purchasing-power-parity (PPP) international dollars. This harmonization reduces differences caused by currencies, price levels, and purchasing power across cities and years. The released field represents the mean property value within a tile in 2017 PPP international dollars. The original property values are also retained for reference and reproducibility. 

Missing values remain unavailable rather than being replaced with zero. The released value remains in physical units; the \texttt{log1p} transformation and train split standardization are applied only in the model input pipeline.

\paragraph{Electricity (ELEC).}
For electricity, we aggregate the valid cell values within each tile by summing them and dividing by the nominal tile area. This produces the tile-level electricity density in kWh/km$^2$/year. For partially supported tiles, we do not estimate the contribution from the missing area. The released value remains in its physical unit, while the model input pipeline applies log transformation and standardization. No additional resampling is applied to the original cell values.

\paragraph{Carbon emissions (ODIAC, CO$_2$).}
For CO$_2$, we aggregate the valid cell values within each tile by summing them and dividing by the nominal tile area. This produces the tile-level fossil-fuel CO$_2$ emission density in tonne C/km$^2$/year. Valid zero values are retained. For partially supported tiles, we do not fill or scale up the missing area. The released value remains in its physical unit, while the model input pipeline applies log transformation and standardization. Each value corresponds to the supplied annual CO$_2$ product, with no additional temporal interpolation.

\paragraph{Nighttime lights (NTL).}
For Nighttime Lights, we compute the masked spatial mean of band 1 radiance within each tile and serialize the result with units of nW/cm$^2$/sr. Radiance is averaged rather than summed because it represents an intensity field. Zero radiance is retained as a valid observation. The released value remains in its physical unit, while the model input pipeline applies log transformation and standardization. No additional sensor harmonization or reconstruction of missing historical years is performed.

\paragraph{GPS trace activity (GPS).}
 For GPS traces, we use tile-level attributes from the OSM GPS Trace GeoJSON rather than a rasterized trajectory image. Records are matched to benchmark tiles using their row and column indices or a parseable tile identifier. A tile is considered unavailable when no corresponding GPS feature or valid point-density value is present. The resulting modality represents the density of GPS sampling points captured in the public OSM trace data. It reflects observed GPS trace activity in the available sample and should not be interpreted as the number of unique travelers or trips, or as a direct measure of overall mobility. 


\subsubsection{Environment modalities Preprocessing}

\paragraph{Optical imagery (OPT).}
  In Table \ref{tab:visual-modalities}, optical imagery is treated as RGB images and retain bands 1-3 in the declared RGB order, while Planet band 4 is excluded. The supplied uint8 RGB values are preserved without per-tile contrast stretching and saved as PNG images. 

\paragraph{Terrain elevation (DEM).}
In Table \ref{tab:visual-modalities}, band 1 of DEM provides elevation in metres. We retain valid zero and negative elevations, exclude pixels marked as missing, and require at least 98\% valid pixels. A fixed global linear grayscale mapping is used to produce DEM PNG image. As a result, a genuinely flat tile may have low visual contrast. DEM is treated as static terrain information and reused across years rather than as an annual measurement, according to previous work \citep{helfenstein2024three}. 

\paragraph{Normalized Difference Vegetation Index(NDVI).}
In Table \ref{tab:visual-modalities}, band 1 of NDVI (Landsat) is rendered as NDVI PNG image, using a fixed palette over the range $[-1,1]$. Zero and negative NDVI values are valid observations rather than nodata. Pixel validity is determined before rendering, and at least 98\% of pixels must be valid. Finite values outside the $[-1,1]$ range are clipped during rendering rather than treated as invalid. We use the annual or composite NDVI provided by the source without recomputing it from the red and NIR bands. No logarithmic transformation or tile-specific contrast stretching is applied.

\paragraph{Air quality (AQ).}
In Table \ref{tab:statistical-modalities}, NO$_2$ and PM$_{2.5}$ are independently aggregated from AirQualityNO2 and AirQualityPM25, respectively. Their coarse source grids are preserved without additional resampling or artificial fine-scale detail. For each product, the tile-level value is computed as the masked spatial mean of valid cells. Valid zero values are retained. If one product is unavailable, it is excluded rather than imputed from the other. The values retain their physical units of ppb for NO$_2$ and $\mu$g/m$^3$ for PM$_{2.5}$, while separate normalizers apply log transformation and standardization. The annual values follow the supplied annual concentration products.

\paragraph{Land-surface temperature (HEAT).}
In Table~\ref{tab:statistical-modalities}, LandsatST uses only bands 1 and 2, which provide the supplied mean and median temperature fields, respectively. We compute the spatial mean of each band. Valid zero degrees Celsius is retained. This annual or composite surface-temperature product is distinct from weather air temperature and is represented as numerical data rather than an image.

\paragraph{Weather (WTH).}
In Table \ref{tab:statistical-modalities}, the weather representation contains TMIN, TMAX, TMEAN, PRCP, SRAD, and RH. For each tile, we first compute an area-weighted spatial
mean over all valid source grid cells. Thus, TMIN, TMAX, and TMEAN are reported
in degrees Celsius, SRAD as mean shortwave radiation, and RH as mean relative
humidity. PRCP is not spatially summed across grid cells.

\section{Any-to-Any Translation}
\label{app:any2any}

\subsection{Experiment Details}

\subsubsection{Any-to-Any Evaluation Metrics}
\label{app:any2any_metrics}

In Table \ref{tab:modus-metrics}, for optical imagery (OPT), we report RGB mean squared error (MSE) and LPIPS to measure pixel-level fidelity and perceptual similarity, respectively.
For continuous spatial fields (DEM, NDVI, and H\&F), we use normalized mean
absolute error (NMAE), while built volume (BV) is evaluated with log1p NMAE
to account for its skewed dynamic range. For categorical land-cover (LC) and
land-use (LU) maps, we use macro mIoU to measure class-wise spatial overlap
while giving equal weight to each class. Road networks (RD) are evaluated
with clDice, which is better suited to thin and connectivity-sensitive
structures.

For reference values $y_i$ and predictions $\hat{y}_i$, we define MSE, NMAE,
and normalized mean squared error (NMSE) as
\begin{equation*}
\mathrm{MSE}
=
\frac{1}{N}\sum_{i=1}^{N}(\hat{y}_i-y_i)^2,
\qquad
\mathrm{NMAE}
=
\frac{\sum_{i=1}^{N}|\hat{y}_i-y_i|}
     {\sum_{i=1}^{N}|y_i|+\epsilon},
\end{equation*}

\begin{equation*}
\tilde{y}_i
=
\frac{T(y_i)-p_{01}}{p_{99}-p_{01}},
\qquad
\tilde{\hat{y}}_i
=
\frac{T(\hat{y}_i)-p_{01}}{p_{99}-p_{01}},
\qquad
\mathrm{NMSE}
=
\frac{1}{N}
\sum_{i=1}^{N}
(\tilde{\hat{y}}_i-\tilde{y}_i)^2.
\end{equation*}

For multi-field numerical targets, including surface temperature (HEAT),
weather (WTH), and air quality (AQ), we first compute NMSE independently
for each target field using its field-specific percentile range, and then
macro-average across the $K$ fields:
\begin{equation*}
\mathrm{macro\ NMSE} = \frac{1}{K}\sum_{k=1}^{K}\left[\frac{1}{N_k}
\sum_{i=1}^{N_k}\left(\tilde{\hat{y}}_{ik}-\tilde{y}_{ik}
\right)^2\right],\end{equation*}
where
\begin{equation*}
\tilde{y}_{ik}=\frac{T_k(y_{ik})-p_{01,k}}     {p_{99,k}-p_{01,k}},
\qquad
\tilde{\hat{y}}_{ik}=\frac{T_k(\hat{y}_{ik})-p_{01,k}}     {p_{99,k}-p_{01,k}}.
\end{equation*}
Here, $p_{01,k}$ and $p_{99,k}$ are the 1st and 99th percentiles of the
training distribution for field $k$. This field-wise normalization prevents
variables with larger numerical ranges from dominating the metric, while
macro-averaging gives equal weight to each target field.

For strongly skewed non-negative targets, we evaluate in
the log-transformed domain by replacing the reference and prediction with
\begin{equation}
y_i'=\log(1+y_i),
\qquad
\hat{y}_i'=\log(1+\hat{y}_i).
\end{equation}
We refer to the resulting metrics as log1p NMAE and log1p NMSE. We use
log1p NMAE for BV and log1p NMSE for CO$_2$, population (POP), property value
(PROP), electricity (ELEC), nighttime lights (NTL), and GPS activity to
reduce the disproportionate influence of extreme values.

For categorical targets, macro mIoU averages IoU equally across semantic
classes, while clDice evaluates centerline agreement and is therefore used
for road-network structure. LPIPS complements MSE for OPT by measuring
similarity in learned perceptual feature space rather than only pixel-wise
differences. For every source-to-target route, the metric is determined
solely by the target modality and is identical across all source modalities.
NMAE and NMSE variants are multiplied by $100$ and reported as percentages;
MSE, LPIPS, macro mIoU, and clDice are reported on their native scales.

\label{app:training-details}

\subsubsection{Benchmark Instance Construction and Data Splits}
\label{app:instance/splits}

\begin{table*}[t]
\centering
\caption{Exhaustive one-to-one candidate pairs for 2-km and 5-km scales (28,249,584 directed pairs). Rows indicate sources and columns indicate targets. Panel (a) lists visual target modalities, and panel (b) lists numerical target modalities.}
\label{tab:cooccurrence}
\scriptsize
\setlength{\tabcolsep}{2.0pt}
\renewcommand{\arraystretch}{1.03}
\begin{minipage}[t]{\textwidth}
\centering
\textbf{(a)}
\vspace{2pt}
\begin{tabular*}{\textwidth}{@{\extracolsep{\fill}}lrrrrrrrrr@{}}
\toprule
Source / Target & OPT & DEM & NDVI & H\&F & BV & LC & LU & POI & RD \\
\midrule
OPT & -- & 193,379 & 203,199 & 178,798 & 207,654 & 167,984 & 119,315 & 166,579 & 159,147 \\
DEM & 193,379 & -- & 120,028 & 104,078 & 123,578 & 102,304 & 72,408 & 85,554 & 88,844 \\
NDVI & 203,199 & 120,028 & -- & 111,913 & 133,144 & 110,430 & 77,667 & 92,491 & 86,312 \\
H\&F & 178,798 & 104,078 & 111,913 & -- & 114,894 & 94,346 & 70,077 & 80,046 & 74,822 \\
BV & 207,654 & 123,578 & 133,144 & 114,894 & -- & 113,390 & 79,632 & 94,950 & 89,072 \\
LC & 167,984 & 102,304 & 110,430 & 94,346 & 113,390 & -- & 72,492 & 74,370 & 69,897 \\
LU & 119,315 & 72,408 & 77,667 & 70,077 & 79,632 & 72,492 & -- & 49,361 & 46,921 \\
POI & 166,579 & 85,554 & 92,491 & 80,046 & 94,950 & 74,370 & 49,361 & -- & 85,117 \\
Road & 159,147 & 88,844 & 86,312 & 74,822 & 89,072 & 69,897 & 46,921 & 85,117 & -- \\
HEAT & 124,092 & 61,789 & 66,897 & 57,811 & 68,575 & 49,580 & 30,376 & 68,939 & 61,945 \\
WTH & 196,112 & 116,784 & 126,187 & 108,757 & 129,457 & 111,090 & 79,737 & 88,757 & 83,029 \\
AQ & 306,563 & 179,484 & 194,021 & 167,699 & 199,163 & 182,159 & 131,631 & 147,200 & 138,905 \\
CO$_2$ & 198,038 & 118,801 & 127,885 & 110,451 & 131,851 & 113,367 & 79,632 & 89,659 & 84,291 \\
POP & 181,718 & 114,530 & 111,106 & 98,774 & 114,530 & 94,300 & 72,387 & 79,290 & 82,232 \\
PROP & 66,712 & 37,008 & 40,986 & 42,245 & 42,063 & 30,137 & 27,809 & 30,544 & 28,347 \\
ELEC & 19,027 & 9,112 & 9,694 & 10,353 & 10,353 & 6,357 & 5,095 & 10,345 & 9,125 \\
NTL & 189,232 & 114,002 & 117,604 & 100,326 & 121,310 & 98,282 & 68,700 & 89,670 & 85,179 \\
GPS & 163,824 & 87,403 & 94,518 & 81,158 & 96,918 & 75,519 & 53,498 & 88,996 & 80,701 \\
\bottomrule
\end{tabular*}
\end{minipage}
\vspace{6pt}
\begin{minipage}[t]{\textwidth}
\centering
\textbf{(b) }
\vspace{2pt}
\begin{tabular*}{\textwidth}{@{\extracolsep{\fill}}lrrrrrrrrr@{}}
\toprule
Source / Target & HEAT & WTH & AQ & CO$_2$ & POP & PROP & ELEC & NTL & GPS \\
\midrule
OPT & 124,092 & 196,112 & 306,563 & 198,038 & 181,718 & 66,712 & 19,027 & 189,232 & 163,824 \\
DEM & 61,789 & 116,784 & 179,484 & 118,801 & 114,530 & 37,008 & 9,112 & 114,002 & 87,403 \\
NDVI & 66,897 & 126,187 & 194,021 & 127,885 & 111,106 & 40,986 & 9,694 & 117,604 & 94,518 \\
H\&F & 57,811 & 108,757 & 167,699 & 110,451 & 98,774 & 42,245 & 10,353 & 100,326 & 81,158 \\
BV & 68,575 & 129,457 & 199,163 & 131,851 & 114,530 & 42,063 & 10,353 & 121,310 & 96,918 \\
LC & 49,580 & 111,090 & 182,159 & 113,367 & 94,300 & 30,137 & 6,357 & 98,282 & 75,519 \\
LU & 30,376 & 79,737 & 131,631 & 79,632 & 72,387 & 27,809 & 5,095 & 68,700 & 53,498 \\
POI & 68,939 & 88,757 & 147,200 & 89,659 & 79,290 & 30,544 & 10,345 & 89,670 & 88,996 \\
RD & 61,945 & 83,029 & 138,905 & 84,291 & 82,232 & 28,347 & 9,125 & 85,179 & 80,701 \\
HEAT & -- & 62,882 & 94,793 & 63,289 & 57,265 & 22,450 & 9,965 & 68,575 & 63,630 \\
WTH & 62,882 & -- & 195,363 & 128,651 & 109,384 & 40,421 & 10,204 & 113,833 & 90,912 \\
AQ & 94,793 & 195,363 & -- & 199,125 & 166,260 & 61,494 & 13,680 & 175,470 & 148,981 \\
CO$_2$ & 63,289 & 128,651 & 199,125 & -- & 110,101 & 40,116 & 10,204 & 116,023 & 91,815 \\
POP & 57,265 & 109,384 & 166,260 & 110,101 & -- & 37,008 & 9,112 & 105,362 & 81,185 \\
PROP & 22,450 & 40,421 & 61,494 & 40,116 & 37,008 & -- & 9,013 & 37,634 & 34,596 \\
ELEC & 9,965 & 10,204 & 13,680 & 10,204 & 9,112 & 9,013 & -- & 10,320 & 10,067 \\
NTL & 68,575 & 113,833 & 175,470 & 116,023 & 105,362 & 37,634 & 10,320 & -- & 89,683 \\
GPS & 63,630 & 90,912 & 148,981 & 91,815 & 81,185 & 34,596 & 10,067 & 89,683 & -- \\
\bottomrule
\end{tabular*}
\end{minipage}
\end{table*}

To prevent spatial leakage across resolutions, training, validation, and test
partitions are defined using a common $10\,\mathrm{km}\times10\,\mathrm{km}$
parent grid. All $2\,\mathrm{km}$ and $5\,\mathrm{km}$ tiles within the same
parent region are assigned to the same partition. All years and modalities associated with these tiles are kept together. Thus, spatially overlapping
observations cannot enter different partitions through a change of scale, modality, or year. Candidate pairs are then sampled independently within each
partition. For categorical modalities, content diversity is considered during
sampling. For example, the coarse spatial support of land use data can produce
tiles containing only one class; more diverse tiles are prioritized to provide
useful supervision, while single-class tiles are retained at a lower sampling
rate rather than removed entirely. Directed source--target routes are assigned
approximately balanced sampling quotas, subject to their available pair counts,
so that frequent routes do not dominate training and rare routes remain
represented. Table~\ref{tab:cooccurrence} shows the number of samples provided by the benchmark. The sample selection procedure produces 269,472 directed translation instances: 172,032 for training, 48,720 for validation, and 48,720 for testing. 

The splits are saved for all models. For each target modality, all source modalities are evaluated on the same target-anchored test instances, identified by identical city, spatial scale, tile, and year keys. The target references are therefore fixed within each column of Table \ref{tab:modus-metrics}, and only the source modality changes. This design makes source-wise performance directly comparable for a fixed target.

\subsubsection{Any-to-any experimental setting.}
All models were fine-tuned on NVIDIA H200 GPUs with 140\,GB of memory. No LoRA modules or parameter-efficient adapters were introduced.
All four baselines are fine-tuned from their released checkpoints without additional urban-specific adapters. For MODUS-7B, we freeze the VAE and fine-tune the LLM and unified diffusion transformer with a learning rate of $1\times10^{-5}$, cosine decay to $5\times10^{-7}$, and a 500-step warm-up on $3\times$ NVIDIA H200 GPUs. For Show-O2-1.5B, we freeze the Wan2.1 VAE and image and positional embeddings, while fine-tuning the remaining trainable components with
learning rates of $5\times10^{-6}$ for the visual encoder and $1\times10^{-5}$ for the projection and generation components, using cosine decay with a 5\% warm-up on one H200 GPU. For UniGen-1.5B, we freeze the MagViT-v2 visual tokenizer and fine-tune the Qwen2.5-1.5B backbone and generation layers with a learning rate of $2\times10^{-5}$, cosine decay to $1\times10^{-6}$, and a 500-step warm-up on one H200 GPU. For ML-4M-3B, we keep the pretrained modality tokenizers frozen and fully fine-tune the 4M transformer with a learning rate of $3.125\times10^{-6}$ and cosine decay on one H200 GPU.

\subsection{Detailed Analysis of Cross-Modal Translation}

\label{app:any2any_detailed_analysis}

\begin{figure}[!htb]
    \centering
    \includegraphics[width=1\linewidth]{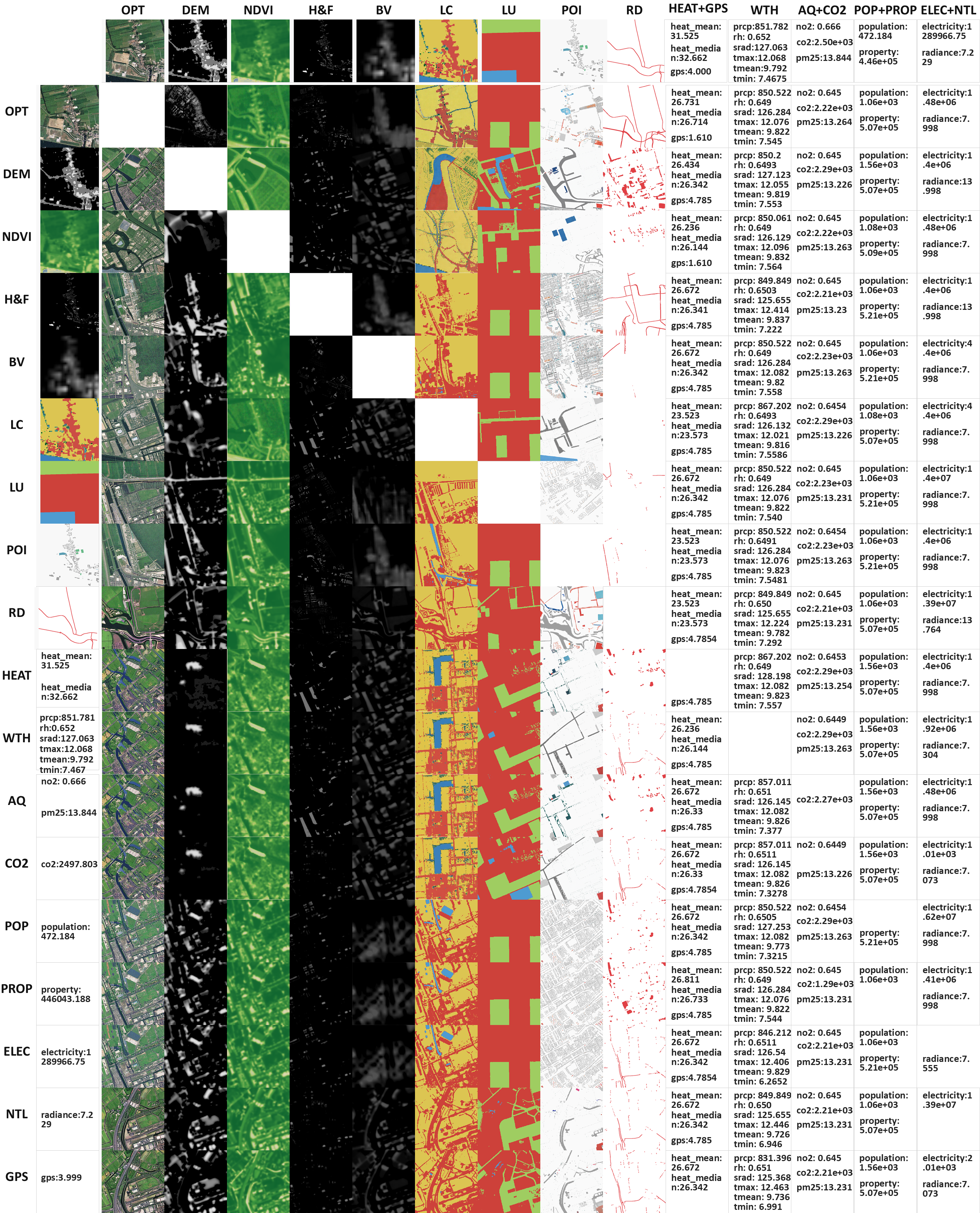}
    \vspace{-1em}
    \caption{Qualitative any-to-any cross-modal translation across 18 urban modalities. Rows denote source modalities and columns denote target modalities. Same-modality source-target pairs are omitted.}
    \vspace{-1em}
    \label{fig:matrix}
\end{figure}

\paragraph{Any-to-any translation reveals cross-modal complementarity and translation relationship.}

This task provides a way to examine how information in one modality can support the generation of another.  Table~\ref{tab:modus-metrics} provides an empirical map of which modalities contain information useful for reconstructing others. Because different targets have different value distributions and prediction characteristics, performance is not directly comparable across different targets. Comparisons are therefore most informative when the target modality is held fixed. Table~\ref{tab:modus-metrics}  shows that activity modalities such as population, electricity, nighttime lights, GPS density, and carbon emissions exhibit clear relationships with each other. For example, GPS$\rightarrow$POP, POP$\rightarrow$ELEC, ELEC$\rightarrow$NTL, and NTL$\rightarrow$GPS achieve NMSEs of 0.219\%, 1.318\%, 2.450\%, and 2.488\%, respectively. For the same target, these activity-related sources also provide useful information compared with other modality types. In particular, POP predicts ELEC with an NMSE of 1.318\%, compared with 37.500\% for RD$\rightarrow$ELEC, indicating that population intensity is considerably more informative about electricity distribution than road geometry alone.

Complementarity also extends across broad modality families. DEM$\rightarrow$WTH obtains 0.756\% NMSE and ranks second among all inputs for WTH; LC$\rightarrow$AQ obtains the best AQ result at 10.153\% NMSE; and OPT$\rightarrow$ELEC achieves the second-best ELEC result at 4.093\% NMSE. These relationships suggest that the model captures a structured dependency graph connecting terrain, land cover, environmental conditions, and human activity, rather than relying only on similarity between modalities of the same type. Nevertheless, these translation relationships should be interpreted as predictive associations rather than causal dependencies.

\paragraph{Cross-modal translation is directionally asymmetric.}
The information that modality $A$ provides about modality $B$ is not generally recoverable in the reverse direction. Because the two directions are evaluated using different target-specific metrics, their raw scores are not directly comparable; nevertheless, their rankings within the corresponding target columns reveal clear asymmetries. For example, DEM$\rightarrow$WTH ranks second among 17 available non-self
inputs, with a macro NMSE of 0.756\%, whereas WTH$\rightarrow$DEM ranks
twelfth, with an NMAE of 25.116\%. Similarly, LC$\rightarrow$AQ is the
best route to AQ, with a macro NMSE of 10.153\%, whereas
AQ$\rightarrow$LC ranks fourteenth, with an mIoU of 0.113. Directionality
is also evident within the numerical modalities:
POP$\rightarrow$PROP is the best route to PROP, with a log1p NMSE of
0.855\%, whereas PROP$\rightarrow$POP is the worst route to POP, with
48.012\%.

The asymmetry is especially pronounced for targets that require accurate
semantic boundaries or topology. Averaged over the available visual and
structural inputs, with unavailable self-translations excluded, the LC,
LU, and RD targets obtain scores of 0.200 mIoU, 0.483 mIoU, and
0.260 clDice, respectively. The corresponding averages for numerical
inputs are 0.137, 0.316, and 0.082. These descriptive gaps suggest that
numerical modalities may constrain the broad appearance, semantic
composition, or overall intensity of an output without uniquely determining
its fine-scale boundaries and topology. This interpretation is also
consistent with Fig.~\ref{fig:matrix}, where numerical inputs produce
globally plausible images and land-cover maps but do not always recover the
ground-truth spatial layout. Conversely, structure-rich inputs preserve
local cues that can be aggregated to predict coarser numerical fields.

\paragraph{Joint any-to-any training is uneven across modalities.}

The imbalance in joint any-to-any training does not arise from uneven sampling. The model must jointly optimize targets that differ not only in semantics, but also in spatial frequency, sparsity, output distribution, and evaluation objective. For example, H\&F and DEM illustrate this conflict particularly clearly. Although both modalities encode height-related information, H\&F is dominated by sparse discontinuities and sharp building contours, whereas DEM primarily contains smooth, low-frequency terrain variation. Shared features and gradients that improve boundary preservation for H\&F can therefore bias DEM generation toward high-frequency structures. In the OPT$\rightarrow$DEM example in Fig.~\ref{fig:matrix}, the generated DEM inherits unnaturally sharp edges and consequently loses part of the smooth terrain pattern. 

More generally, joint training must reconcile pixel-wise regression and perceptual reconstruction for continuous images, semantic objectives for LC and LU, and a topology-sensitive objective for RD. Improvements induced by one type of supervision therefore need not transfer uniformly to the others. The results in Table~\ref{tab:5route}, where different models perform best on different translation routes rather than one model consistently dominating all targets, are consistent with such cross-task interference. Joint training consequently provides modality-dependent benefits rather than a uniform improvement across the translation matrix. 

These observations motivate a combination of shared and modality-specific optimization. Possible strategies include route-balanced sampling, target-dependent loss normalization or gradient balancing, modality-specific normalization layers, and lightweight output adapters. Frequency-aware objectives may be particularly useful for the height modalities: edge-sensitive supervision can preserve building boundaries in H\&F, while smoothness or low-frequency consistency constraints can prevent the same edge prior from leaking into DEM generation.

\section{Agentic Tasks}
\label{app:agentic_tasks}
\subsection{Case Construction and Selection}
\label{app:agentic-case-construction}

The case counts and geographic coverage are reported in the main text. Here we describe the eligibility criteria, sampling procedures, and reference outcomes used to construct the evaluation set. Selection operates on task-specific records linked to the benchmark grid: a tile--target pair for forecasting, a tile for EV expansion and food access, and an individual school for capacity planning. Reference outcomes are stored separately from agent-visible observations.

\paragraph{Forecasting.}
The forecasting targets use the benchmark's annual Landsat surface temperature products \citep{usgs_landsat_c2_st}, ODIAC carbon-emission (CO$_2$) records \citep{odiac_data_product}, and city-specific property records documented in Appendix~B.2. For each city and target, we retain tiles with valid annual target observations throughout 2010--2024 and a consistent measurement unit across years. This complete-panel requirement supports all five forecast origins on the same spatial sample. The initial cohort samples two eligible tiles per city--target combination using a fixed pseudorandom seed of 0. The expanded cohort adds 14 LST, 12 carbon-emission, and 16 property-price tile--target pairs, excluding those already selected for the same target. Additional candidates are ordered by hash with seed \texttt{prediction600-seed42} and selected in round-robin order across eligible cities. Selection depends on data availability and city coverage, without ranking candidates by target magnitude. Each selected pair generates five cases; the reference answer is the observed target value at the corresponding forecast origin. This yields 44 LST, 44 carbon-emission, and 32 property-price spatial series. Different targets may share a tile, while each tile--target--year combination remains a distinct case.

\paragraph{EV charging expansion.}
We use the archived alternative-fueling-station records underlying the EV case package, associated with the Alternative Fuels Data Center station database \citep{doe_afdc_stations}. We construct reference outcomes by aggregating public charging-port counts to tiles at the baseline and target dates and taking their difference. These observed changes define four expansion categories using city-specific intervals (Table~\ref{tab:ev-case-strata}). Sampling is stratified jointly by city and category. Within each stratum, candidates are ordered by a deterministic hash of the seed, city, category, and tile identifier. Pilot tiles are excluded, and the additional cohort also excludes previously selected tiles. The final sample contains 38 cases in each of categories A and B and 37 in each of C and D. The category boundaries are derived from the full observed outcome data; the resulting sample provides approximately balanced coverage of expansion levels rather than their population frequencies.

\begin{table}[!htbp]
\centering
\small
\caption{City-specific strata for EV case selection, defined by the observed net increase in public charging ports over the evaluation horizon.}
\label{tab:ev-case-strata}
\begin{tabular}{lrrrr}
\hline
City & A & B & C & D \\
\hline
Boston & $\leq 0$ & $1$--$5$ & $6$--$10$ & $\geq 11$ \\
New York City & $\leq 0$ & $1$--$4$ & $5$--$18$ & $\geq 19$ \\
San Francisco & $\leq 0$ & $1$--$3$ & $4$--$8$ & $\geq 9$ \\
\hline
\end{tabular}
\end{table}

\paragraph{School capacity planning.}
We use the UK Department for Education's \emph{School capacity: Academic year 2024/25} release, which provides school-level places and pupil enrollment records \citep{dfe_school_capacity_202425}. School locations are joined using coordinates derived from Get Information about Schools \citep{dfe_gias}. We link longitudinal records by the school Unique Reference Number (URN) and retain primary and secondary schools with exactly one record in each required academic year, a consistent education phase, valid nonnegative enrollment, positive capacity, and a coordinate that maps to the benchmark grid. Schools are assigned to development or test pools by a deterministic tile-level hash, keeping colocated schools in the same pool. Evaluation cases are selected from the test pool with at most one school per tile. Let $E_{+1}$ denote observed target-year enrollment and $C_0$ the last historical capacity. Sampling distinguishes balanced cases with $0.95 \leq E_{+1}/C_0 \leq 1.05$ from shortage cases with $E_{+1}/C_0 > 1.05$; surplus cases are excluded. The final cohort retains the 20 pilot cases and fills the two strata to 50 cases each using a fixed hash ordering. Target-year enrollment supplies the reference demand for evaluating the proposed capacity, while the surrounding tile supplies contextual information for the individual school.

\paragraph{Food-access analysis.}
We derive reference labels from the 2019 Food Access Research Atlas of the USDA Economic Research Service \citep{usda_fara_2019}. The source provides census-tract-level low-income and food-access indicators under the urban 1-mile and rural 10-mile distance convention. These tract indicators are spatially aligned to the four assessment cells within each tile. The resulting cell labels distinguish low-income-and-low-access, low-access without low-income status, and not-low-access areas. Source coverage and assignment scores are retained alongside each label. Eligible tiles must contain all four assessment cells, each with at least 95\% source coverage and a valid label. Tiles are stratified by the number of cells marked as low access, including those also classified as low income: zero, one to three, or four. Within each city and stratum, selection prioritizes the number of low-income-and-low-access cells, then spatial separation from already selected tiles measured by Manhattan distance on the grid, then the minimum source-assignment score across the four cells; a deterministic hash resolves ties. Pilot tiles are excluded, and the expansion excludes the original cohort. The final set contains 48 tiles with no low-access cells, 54 with mixed access, and 48 with all four cells marked as low access. This procedure preserves varied within-tile patterns and emphasizes underserved areas; these strata describe sampling composition rather than citywide prevalence.

\subsection{Experimental Setup}
\paragraph{Agent tools.}
Tool access is defined separately for each task.
For forecasting, Earth Agent uses specialized statistical tools,
including trend estimation, summary statistics, and time-series
diagnostics, together with tools for retrieving target histories,
discovering assets, and inspecting tables and rasters.
Its adapter retains 90 of the 104 tools in the original implementation
and exposes 20 additional helper functions. The excluded tools require
unavailable model assets or GDAL-specific processing.
Codex and OpenCode retain their native shell and file-operation tools
within case-scoped workspaces, allowing them to implement analyses through code. Forecasting comparisons therefore evaluate complete agent configurations with different analytical interfaces.

For planning and resource analysis, all configurations share the task-specific tool interface within each task.
EV charging provides nine tools for retrieving charging histories, discovering and inspecting urban data, calculation, and submission. School capacity planning provides eight tools for retrieving school histories, inspecting urban assets, calculation, and plan submission.
Food-access analysis provides seven tools for asset discovery,
table and raster inspection, cell-level summaries, calculation,
and submission of the four cell labels.
Tools expose source metadata and observation periods;
reference answers and target-period outcomes are unavailable
through these interfaces. A successful task-specific submission
terminates the case.

Agents choose which available tools to invoke and which data
to inspect. Tool-call counts describe interactions at each framework's
interface rather than equivalent amounts of computation: a single
shell invocation may perform multiple reads and calculations.
The counting protocol and observed tool-use statistics are reported
in Appendix~\ref{app:agent-tool-use}.
\paragraph{Agentic evaluation metrics.}
We use task-specific metrics to evaluate forecasting, urban planning, and resource analysis.
\textbf{(1) Forecasting:} Land-surface-temperature and carbon-emission forecasts are evaluated using mean absolute error (MAE), measured in $^\circ$C and tonnes C/tile/year, respectively. Property-price forecasts are evaluated using mean absolute percentage error (MAPE) to accommodate different local currencies:
\[
\mathrm{MAE}=\frac{1}{N}\sum_{i=1}^{N}|\hat{y}_i-y_i|,
\qquad
\mathrm{MAPE}=\frac{100}{N}\sum_{i=1}^{N}
\frac{|\hat{y}_i-y_i|}{|y_i|},
\]
where $y_i$ and $\hat{y}_i$ denote observed and predicted values. We report results separately for each forecasting target.

\textbf{(2) EV charging:} We evaluate predictions of the four charging-port expansion categories using accuracy, macro-F1, and balanced accuracy. Macro-F1 averages class-specific F1 scores, while balanced accuracy averages class-specific recall.

\textbf{(3) School capacity:} We evaluate next-year enrollment forecasts using MAE and capacity plans using the balanced-case rate and mean capacity-balance score. Let $E$ denote observed target-year enrollment and $C=C_0+\Delta C$ denote existing capacity plus proposed additional seats. A plan is balanced when $0.95\leq E/C\leq1.05$, and its capacity-balance score is
\[
S_{\mathrm{balance}}
=100\left(1-\frac{|C-E|}{C}\right).
\]
Higher scores indicate closer alignment between capacity and observed enrollment. We additionally report total added seats, shortfall seats $\max(0,E-C)$, and surplus seats $\max(0,C-E)$, separating pre-existing surplus from surplus introduced by expansion.

\textbf{(4) Food accessibility:} We report cell-level accuracy and macro-F1 across the three accessibility categories over 600 assessment cells, together with the number of tiles for which all four cells are classified correctly. These metrics measure agreement with the benchmark's tract-derived reference labels.

\subsection{Task Setting}
\subsubsection{Task Instruction}
\label{app:agent-prompts}
We report the case-independent task instructions used in the six evaluated tasks, formatted for the appendix. Case IDs, locations, historical observations, available assets, and answer ranges are supplied by the corresponding case or tools. The three forecasting targets share one instruction template and differ in the target fields of the case message. EV charging, school capacity, and food access use separate task prompts. 

\paragraph{Forecasting: land surface temperature, carbon emissions, and property values.}
The shared forecasting instruction reads:
\begin{quotation}\small
Predict the requested annual value for one $2\,\mathrm{km}\times2\,\mathrm{km}$ tile at period $0$, using the supplied ten-year history (periods $-10$ to $-1$). The case message specifies the target, unit, and spatial aggregation. Calendar years are hidden; no target-period observations are supplied. Available modalities include satellite imagery, surface temperature, vegetation, land cover, terrain, buildings, weather, population, nighttime lights, carbon emissions, and property prices; availability varies by tile and period. You may explore multiple relevant modalities, including imagery, to inform your prediction. Use \texttt{get\_history} for the target history and \texttt{list\_assets} to discover available modalities. Use \texttt{read\_table} for numerical properties, \texttt{describe\_raster} for bands and statistics, \texttt{view\_raster} for image previews, and \texttt{materialize\_asset} when an analysis tool needs a local asset path. Return one JSON object with \texttt{prediction} (number), \texttt{unit}, \texttt{reason}, \texttt{modalities\_used}, and \texttt{modules\_used}; list only sources and tools actually used.
\end{quotation}
The case message supplies the city and tile, \texttt{prediction\_period=0}, the target, its unit and aggregation, and the history window $[-10,-1]$. The three task-specific instantiations are:
\begin{center}
\small
\begin{tabular}{lll}
\toprule
Task & Target field & Requested output \\
\midrule
LST & \texttt{lst} & Period-$0$ tile-mean LST in $^\circ\mathrm{C}$ \\
Carbon emissions & \texttt{carbon} & Period-$0$ tile total in tonnes C/tile/year \\
Property values & \texttt{property} & Period-$0$ median of valid 500-m cell medians \\
\bottomrule
\end{tabular}
\end{center}
In each case, the ten historical values are retrieved through \texttt{get\_history}; the target-period value is withheld. Property values use the city's local currency.

\paragraph{EV charging-site planning.}
The task prompt is instantiated with the case ID, city, tile, and four city-specific port-growth ranges:
\begin{quotation}\small
Forecast the net change in publicly available EV charging ports in this $2\,\mathrm{km}\times2\,\mathrm{km}$ tile from $t-2$ to $t$ (two years). Choose one of the supplied ranges A--D for $\mathrm{ports}(t)-\mathrm{ports}(t-2)$. Use \texttt{inspect\_case} for charging history at $t-4$, $t-3$, and $t-2$. Use \texttt{list\_modalities} and \texttt{list\_assets} to discover historical data through $t-2$; inspect relevant tables, rasters, and baseline records, and use \texttt{calculate} for arithmetic. Each time unit is one year; calendar years are hidden. Ports count vehicles that can charge simultaneously (Level 1, Level 2, and DC fast), whereas stations are separate counts. Option A includes zero or negative growth. Follow each source's units and observation dates; missing values are unknown, not zero. Use only this case's supplied tools and evidence. Do not access future data, answer labels, other cases, or the web. Cite only tool-call IDs marked citable and distinguish observations from inference. Call \texttt{submit\_forecast} with the selected range, confidence, a concise explanation, a reason for each A--D range, evidence-call IDs, and limitations. A successful submission ends the task.
\end{quotation}

\paragraph{School capacity planning.}
The school prompt is instantiated with a school case ID and city:
\begin{quotation}\small
Forecast next-year enrollment and plan period-$+1$ capacity for the specified school. Read enrollment and capacity for periods $-2$, $-1$, and $0$ with \texttt{read\_school\_history}. Choose an integer \texttt{added\_seats}$\geq 0$; planned capacity equals period-$0$ capacity plus added seats. Reductions are unavailable, but zero additions are allowed. Use only supplied case tools and evidence through period $0$. Urban evidence is optional: check its age, units, and spatial coverage. A tile is not a verified school catchment, and a single-period proxy cannot establish growth. Missing data are unknown; do not access future outcomes or infer student counts from display pixels. Aim for next-year enrollment to be 95\%--105\% of planned capacity. Submit \texttt{predicted\_enrollment}, \texttt{added\_seats}, evidence, and a concise reason through \texttt{submit\_plan}. Explain the forecast calculation and assumptions, compare utilization before and after expansion, and justify additions or zero. Cite each factual basis using a successfully read \texttt{school-history} or asset reference, separating observations from inference. A successful submission ends the task.
\end{quotation}

\paragraph{Food access.}
The food-access prompt is instantiated with a tile case ID and city:
\begin{quotation}\small
Predict FoodAccess for four $1\,\mathrm{km}\times1\,\mathrm{km}$ cells in this $2\,\mathrm{km}\times2\,\mathrm{km}$ tile: \texttt{s00} (NW), \texttt{s01} (NE), \texttt{s10} (SW), and \texttt{s11} (SE). Assign each cell one of \texttt{not\_low\_access\_1and10}, \texttt{low\_access\_1and10}, or \texttt{lila\_1and10}. Low access means that at least 500 people or 33\% of a census tract live beyond one mile (urban) or ten miles (rural) from a qualifying supermarket; these tract-derived labels are transferred to the grid rather than measured independently per cell. Discover assets with \texttt{list\_reference\_assets}; inspect relevant tables, raster descriptions, images, and cell-level summaries. Read units, class legends, source ages, and spatial support. Worksite earnings and property values are indirect income evidence; generic retail or food-service classes are not supermarkets. Missing evidence is unknown, not absence. Assess access and income separately for each cell, distinguishing observations from inference. Use only this case's tools and evidence, without labels, other cases, web, filesystem, or future data. Call \texttt{submit\_predictions} with one label, confidence, reason, and evidence for each cell, plus limitations. Each cited observation must refer to a successfully inspected result in the form \texttt{evidence-NNNN/asset\_ID: observation}. A successful submission ends the task.
\end{quotation}

\subsubsection{Example Task Cases}
\label{app:prediction-case-study}

The four forecasting examples below pair held-out reference outcomes with submitted answers and verbatim excerpts from the agents' final rationales. All four are Codex (DeepSeek) cases from the 600-case forecasting evaluation. Reference outcomes and errors were added by the evaluator and were withheld from the agent. These cases illustrate individual decisions rather than aggregate performance.
\begin{table}[H]
\centering
\small
\caption{LST forecasting: retaining the historical mean after inspecting auxiliary modalities.}
\label{tab:lst-case}
\renewcommand{\arraystretch}{1.12}
\begin{tabularx}{\textwidth}{@{}X@{}}
\toprule
\textbf{Task} \\[2pt]
Predict period-$0$ annual tile-mean land-surface temperature from ten historical periods and available urban modalities. \\
\midrule
\textbf{Case and reference outcome} \\[2pt]
\textbf{Case ID:} \texttt{Boston\_2KM\_r7\_c8\_\_lst\_\_2022} \\
\textbf{City:} Boston. \textbf{Forecast period:} $0$; calendar year withheld. \\
\textbf{Reference:} 19.910 $^{\circ}$C. \textbf{Last year:} 21.990. \textbf{Ten-year mean:} 21.310. \\
\midrule
\textbf{Submitted answer and evaluated error} \\[2pt]
\textbf{Codex (DeepSeek):} 21.310 $^{\circ}$C. \\
Absolute error: 1.400 $^{\circ}$C. Last-year baseline error: 2.080; ten-year-mean baseline error: 1.400. \\
\midrule
\textbf{Agent's submitted rationale (verbatim excerpts)} \\[2pt]
Weather, vegetation, air-quality, night-light, land-cover, and imagery layers were inspected as possible predictors. They are contemporaneous historical covariates only; no period-0 weather or other future covariate is supplied. Regressions using them were unstable on only ten annual observations and were therefore not allowed to override the target series.
\par\medskip\textnormal{[\ldots]}\par\medskip
Its mean is \texttt{21.30999984741211 degC}. The fitted linear-trend slope is not statistically distinguishable from zero, and the AR(1) coefficient is also weak. A rolling one-step-ahead check from four forecast origins gave the lowest mean absolute error to the level rule; its rolling errors were smaller than those of persistence and comparable to, but more stable than, fitted trend or autoregressive rules.
\par\smallskip
Therefore the period-0 forecast is the level expectation of the ten supplied annual values:
\par\smallskip
forecast = mean(period -10 through -1) = 21.31 degC \\
\bottomrule
\end{tabularx}
\end{table}

\begin{table}[H]
\centering
\small
\caption{Carbon forecasting: auxiliary histories inform the treatment of a temporary shock.}
\label{tab:carbon-case}
\renewcommand{\arraystretch}{1.12}
\begin{tabularx}{\textwidth}{@{}X@{}}
\toprule
\textbf{Task} \\[2pt]
Predict period-$0$ annual carbon emissions summed over the four complete 1 km support cells in a 2 km tile, using ten historical periods and available urban modalities. \\
\midrule
\textbf{Case and reference outcome} \\[2pt]
\textbf{Case ID:} \texttt{London\_2KM\_r20\_c14\_\_carbon\_\_2022} \\
\textbf{City:} London. \textbf{Forecast period:} $0$; calendar year withheld. \\
\textbf{Reference:} 1,980.968 tonnes C/tile/year. \textbf{Last year:} 2,053.225. \textbf{Ten-year mean:} 2,430.620. \\
\midrule
\textbf{Submitted answer and evaluated error} \\[2pt]
\textbf{Codex (DeepSeek):} 1,965.800 tonnes C/tile/year. \\
Absolute error: 15.168 tonnes C/tile/year. Last-year baseline error: 72.258; ten-year-mean baseline error: 449.652. \\
\midrule
\textbf{Agent's submitted rationale (verbatim excerpts)} \\[2pt]
Period -2 is a transient shock rather than the normal trend: carbon falls 9.63\% from period -3 and rebounds 4.49\% in period -1. The same pattern is present in independently supplied PM2.5 and nighttime-light histories, while residential electricity rises in period -2 and falls in period -1.
\par\smallskip
To avoid fitting the shock, I estimated the ordinary geometric decline from the seven pre-shock transitions, periods -10 through -3. The mean log change is \texttt{-0.0435127711593406} per period. Applying that rate once to the period -1 target gives:
\par\smallskip
\texttt{2053.2254638671875 * exp(-0.0435127711593406) = 1965.799794292253}. \\
\bottomrule
\end{tabularx}
\end{table}

\begin{table}[H]
\centering
\small
\caption{Carbon forecasting: inspected auxiliary modalities do not alter the point forecast.}
\label{tab:carbon-case2}
\renewcommand{\arraystretch}{1.12}
\begin{tabularx}{\textwidth}{@{}X@{}}
\toprule
\textbf{Task} \\[2pt]
Predict period-$0$ annual carbon emissions summed over the four complete 1 km support cells in a 2 km tile, using ten historical periods and available urban modalities. \\
\midrule
\textbf{Case and reference outcome} \\[2pt]
\textbf{Case ID:} \texttt{Amsterdam\_2KM\_r7\_c11\_\_carbon\_\_2021} \\
\textbf{City:} Amsterdam. \textbf{Forecast period:} $0$; calendar year withheld. \\
\textbf{Reference:} 1,759.693 tonnes C/tile/year. \textbf{Last year:} 1,772.050. \textbf{Ten-year mean:} 1,963.619. \\
\midrule
\textbf{Submitted answer and evaluated error} \\[2pt]
\textbf{Codex (DeepSeek):} 1,728.772 tonnes C/tile/year. \\
Absolute error: 30.921 tonnes C/tile/year. Last-year baseline error: 12.357; ten-year-mean baseline error: 203.926. \\
\midrule
\textbf{Agent's submitted rationale (verbatim excerpts)} \\[2pt]
Historical nighttime lights, electricity use, gas use, population, GDP, built volume, weather, and imagery were inspected as possible secondary modalities. They do not contain period-0 observations and have less complete or more indirect relationships to this target, so they were not used to alter the point forecast.
\par\medskip\textnormal{[\ldots]}\par\medskip
The forecast uses an unweighted log-linear trend over periods \texttt{t = -10, ..., -1}:
\par\smallskip
log(carbon\_t) = a + b t
\par\smallskip
The fitted coefficients are:
\par\smallskip
a = 7.455166557121871 b = -0.0226900867302732
\par\smallskip
Thus the fitted annual change is \texttt{exp(b) - 1 = -2.2434602679\%}, and the period-0 forecast is:
\par\smallskip
carbon\_0 = exp(a) = 1728.7719105633914
\par\smallskip
This gives \texttt{1728.771910563391} after normal decimal rounding. \\
\bottomrule
\end{tabularx}
\end{table}

\begin{table}[H]
\centering
\small
\caption{Property forecasting: distinguishing a transformed target series from independent evidence.}
\label{tab:property-case}
\renewcommand{\arraystretch}{1.12}
\begin{tabularx}{\textwidth}{@{}X@{}}
\toprule
\textbf{Task} \\[2pt]
Predict the period-$0$ property value of a 2 km London tile: the median of valid 500 m cell-level property-value medians, using ten historical periods and available urban modalities. \\
\midrule
\textbf{Case and reference outcome} \\[2pt]
\textbf{Case ID:} \texttt{London\_2KM\_r16\_c26\_\_property\_\_2021} \\
\textbf{City:} London. \textbf{Forecast period:} $0$; calendar year withheld. \\
\textbf{Reference:} 390,000.000 GBP. \textbf{Last year:} 347,500.000. \textbf{Ten-year mean:} 303,337.375. \\
\midrule
\textbf{Submitted answer and evaluated error} \\[2pt]
\textbf{Codex (DeepSeek):} 347,500.000 GBP. \\
Absolute percentage error: 10.90\%. Last-year baseline: 10.90\%; ten-year-mean baseline: 22.22\%. \\
\midrule
\textbf{Agent's submitted rationale (verbatim excerpts)} \\[2pt]
I also examined the supplied property distribution details, the PPP/GDP-adjusted property series, nighttime lights, GDP, built volume, electricity and gas use, population, vegetation, and OSM infrastructure. These modalities were useful as consistency checks but could not supply a direct period-0 observation or a validated cross-modal improvement over the local property history.
\par\medskip\textnormal{[\ldots]}\par\medskip
The forecast is the median of four robust local-level estimates:
\par\smallskip
1. the most recent annual value: 347500.0; 2. the median of the last three annual values: 347500.0; 3. an AR(1) one-step forecast: 345222.92; 4. an exponentially weighted forecast with alpha 0.7: 349231.60.
\par\smallskip
The median of these four candidates is:
\par\smallskip
period-0 property value = 347500.0 GBP
\par\medskip\textnormal{[\ldots]}\par\medskip
The PPP/GDP-adjusted property series is effectively a period-specific scaling of the same underlying property observations. Projecting its recent scale ratio implies approximately 343228 GBP, within about 1.2\% of the selected forecast. The property mean plus a recent median-minus-mean spread gives approximately 346410 GBP, also close to the selected result. \\
\bottomrule
\end{tabularx}
\end{table}

\ifdefined\FloatBarrier\FloatBarrier\fi

Together, these four cases illustrate different roles for auxiliary evidence in forecasting. In the LST case, the agent retains the historical mean after finding unstable relationships with auxiliary variables, outperforming the
last-year baseline. In the London carbon case, cross-modal observations inform the identification of a temporary shock and the choice of trend window, producing a forecast closer to the reference than either historical baseline. In contrast, the Amsterdam carbon case uses auxiliary modalities only as checks, and its trend forecast performs worse than persistence. The property
case correctly identifies a transformed target series as non-independent evidence, yet its final estimate remains equal to the last-year value and misses the subsequent increase. These cases show that recognizing relevant
or redundant evidence does not by itself ensure an accurate forecast; the challenge is deciding whether, and how, that evidence should change the forecasting rule.
Together, the four forecasting cases show how auxiliary evidence enters model selection, trend-window selection, and consistency checks. For Boston LST, the agent evaluates auxiliary regressions and retains the ten-year mean, reducing absolute error from 2.080 to 1.400 $^{\circ}$C relative to the last-year baseline. For London carbon emissions, concurrent changes in PM$_{2.5}$, nighttime lights, and electricity use inform the treatment of a temporary shock and the choice of a pre-shock trend window; absolute error falls from 72.258 to 15.168 tonnes C/tile/year. For Amsterdam carbon emissions, the agent inspects auxiliary modalities and selects a historical log-linear trend, yielding an absolute error of 30.921 compared with 12.357 for the last-year baseline. For London property values, the agent identifies the PPP/GDP-adjusted series as a transformation of the target and selects a local-level forecast with a 10.90\% absolute percentage error. These decisions highlight the importance of connecting modality semantics and temporal evidence to the choice and calibration of a forecasting rule.

The following three examples examine infrastructure-growth reasoning, capacity margins, and service-specific spatial interpretation. The EV case links an underestimated expansion to an assumption that a recent installation wave has temporarily satisfied local opportunities. The school case shows how a small enrollment underestimate affects a plan placed near the upper utilization threshold. The food-access case shows how general commercial activity enters a supermarket-access judgment after the agent has identified the limitations of the POI categories.

\begin{table}[!htbp]
\centering\small
\caption{EV charging-port expansion: a recent buildout is interpreted as temporary saturation.}
\label{tab:ev-case}
\renewcommand{\arraystretch}{1.10}
\begin{tabularx}{\textwidth}{@{}X@{}}
\toprule

\textbf{Task and case} \\[2pt]
Forecast the net change in public charging ports over two years and select a city-specific growth range. \\
\textbf{Case ID:} \texttt{evport\_494be8490c9337a6}. \textbf{Tile:} \texttt{Boston\_2KM\_r6\_c3}. \\
\textbf{Options:} A: $\leq0$; B: 1--5; C: 6--10; D: $\geq11$ additional ports. \\
\midrule
\textbf{Submitted answer and observed outcome} \\[2pt]
\textbf{Codex (GPT-5.5):} B (1--5 ports), confidence 0.38. \\
\textbf{Reference:} D; the observed count rises from 21 to 39 ports, a net increase of 18. \\
\midrule
\textbf{Agent's submitted rationale (verbatim excerpt)} \\[2pt]
The tile is dense and urban, with high population/POI/building intensity and substantial electricity use, supporting some continued demand, but the recent 21-port buildout may have temporarily satisfied local opportunities, making another $\geq$6-port increase less certain over the next two years. \\
\midrule
\textbf{Case analysis} \\[2pt]
The agent recognizes population, POI, building-intensity, and electricity-use signals. Its final category follows the assumption that the recent 21-port buildout has temporarily satisfied local opportunities. The observed expansion shows the importance of evaluating that saturation assumption when translating urban activity into infrastructure-growth forecasts. \\

\bottomrule
\end{tabularx}
\end{table}

\begin{table}[!htbp]
\centering\small
\caption{School capacity planning: a narrow capacity margin is sensitive to enrollment error.}
\label{tab:school-case}
\renewcommand{\arraystretch}{1.10}
\begin{tabularx}{\textwidth}{@{}X@{}}
\toprule

\textbf{Task and case} \\[2pt]
Forecast next-period enrollment and choose nonnegative additional seats to achieve enrollment-to-capacity utilization of 95--105\%. \\
\textbf{Case ID:} \texttt{school\_plan\_d2d749da062a}. \textbf{Tile:} \texttt{London\_2KM\_r10\_c20}. \\
Historical enrollment is 637, 724, and 753; existing capacity is 600 seats. \\
\midrule
\textbf{Submitted answer and observed outcome} \\[2pt]
\textbf{Earth Agent (DeepSeek):} forecast 782 pupils; add 145 seats; planned capacity 745. Forecast utilization is 104.966\%. \\
\textbf{Reference enrollment:} 788 pupils. Forecast error is 6 pupils; observed utilization is 105.772\%, exceeding the upper threshold. \\
\midrule
\textbf{Agent's submitted rationale (verbatim excerpts)} \\[2pt]
I take the most recent increment (+29) as the conservative near-term trend and forecast period +1 enrollment = 753 + 29 = 782.
\par\medskip\textnormal{[\ldots]}\par\medskip
Forecast utilization after expansion = 782/745 = 105.0\%, the top of the target band; it also keeps the plan at the minimum expansion consistent with the band, avoiding over-provisioning. \\
\midrule
\textbf{Case analysis} \\[2pt]
The minimum-expansion decision leaves approximately 0.034 percentage points below the forecast-utilization ceiling. The six-pupil forecast error moves observed utilization above the target band. The agent also calculates a candidate capacity of 760 seats; evaluated against observed enrollment, this candidate gives 103.684\% utilization. The case identifies capacity-margin selection as a concrete link between enrollment uncertainty and expansion decisions. \\

\bottomrule
\end{tabularx}
\end{table}

\begin{table}[!htbp]
\centering\small
\caption{Food-access classification: commercial activity is used to infer supermarket accessibility.}
\label{tab:foodaccess-case}
\renewcommand{\arraystretch}{1.10}
\begin{tabularx}{\textwidth}{@{}X@{}}
\toprule

\textbf{Task and case} \\[2pt]
Classify the four 1 km cells of a 2 km tile by food access and low-income status. \\
\textbf{Case ID:} \texttt{fa\_full\_221fd43d1213}. \textbf{Tile:} \texttt{DC\_2KM\_r8\_c8}. \\
\textbf{Labels:} NLA: not low access; LA: low access; LILA: low income and low access. Reference labels are transferred from USDA census-tract indicators to the assessment cells. \\
\midrule
\textbf{Submitted answer and reference labels} \\[2pt]
\textbf{Earth Agent (DeepSeek):} NLA for NW, NE, SW, and SE. \\
\textbf{Reference:} NW, SW, and SE are LILA; NE is NLA. One of four labels is correct. \\
\midrule
\textbf{Agent's submitted rationale (verbatim excerpts)} \\[2pt]
\textit{Recorded limitation:} No asset identifies qualifying supermarkets: OSM\_POI/OSM\_Amenity legends explicitly state food\_service and commercial\_retail\_office are not supermarkets, so supermarket proximity (the core of the low-access definition) is unobserved for every cell.
\par\medskip
\textit{NW-cell judgment:} Dense urban DC with the richest retail/POI mix makes supermarket access within 1 mile most likely; no income evidence supports a lila classification. \\
\midrule
\textbf{Case analysis} \\[2pt]
The agent identifies the semantic gap between food-service or commercial categories and qualifying supermarkets. Its final judgment nevertheless uses retail intensity to infer supermarket access. The observed errors connect this proxy interpretation to the classification outcome. The case motivates stronger checks on service identity, distance support, and income evidence when converting spatial observations into cell labels. \\

\bottomrule
\end{tabularx}
\end{table}

\FloatBarrier
\subsection{Task Result}
\label{app:agentic-task-results}

\paragraph{Additional forecasting metrics}
\label{app:prediction-additional-metrics}
Tables~\ref{tab:appendix-lst-rmse-r2}--\ref{tab:appendix-property-rmse-r2-4} supplement MAE and MAPE with RMSE and $R^2$:
\begin{equation}
\mathrm{RMSE}=\sqrt{\frac{1}{N}\sum_{i=1}^{N}(y_i-\hat y_i)^2},\qquad
R^2=1-\frac{\sum_{i=1}^{N}(y_i-\hat y_i)^2}{\sum_{i=1}^{N}(y_i-\bar y)^2}.
\end{equation}
Here, $\bar y$ is the reference mean in the evaluated group. LST and carbon metrics pool the cases within each task. Property metrics are computed separately by city in local currency. Pooled $R^2$ summarizes agreement across the spatial and temporal variation represented in the evaluation cohort.

\paragraph{Tile-block bootstrap.}
Bracketed values give pointwise 95\% percentile intervals from 10,000 city-stratified tile-block bootstrap replicates, using random seed 20261024. A block contains all five forecast years of one tile. For each city, tiles are sampled with replacement while retaining its original tile count and all five observations per sampled tile. Each replicate recomputes RMSE, the reference mean, and $R^2$. All reported intervals use the same sampled blocks for every method within an evaluation group. LST and carbon each contain 44 blocks; each property city contains four. The intervals describe sensitivity to the sampled spatial tiles, conditional on the observed cities and years. Each property city contributes 20 forecasts from four spatial series, so its interval is particularly sensitive to which tiles are sampled. Each entry reports a point estimate followed by its interval.

\begin{table}[!htbp]
\centering\small
\caption{Land-surface temperature: point estimates and 95\% tile-block bootstrap intervals on 220 cases. RMSE is in $^\circ$C.}
\label{tab:appendix-lst-rmse-r2}
\begin{adjustbox}{max width=\textwidth}
\begin{tabular}{lrr}
\toprule
Method & RMSE $\downarrow$ [95\% CI] & $R^2\uparrow$ [95\% CI] \\
\midrule
Earth Agent (GPT Luna) & 1.778 [1.582, 1.974] & 0.899 [0.866, 0.924] \\
Earth Agent (DeepSeek) & 1.761 [1.510, 2.021] & 0.901 [0.860, 0.930] \\
Codex (GPT-5.5) & 1.671 [1.498, 1.847] & 0.910 [0.884, 0.931] \\
Codex (DeepSeek) & 1.705 [1.527, 1.883] & 0.907 [0.878, 0.928] \\
OpenCode (GPT Luna) & 1.677 [1.483, 1.869] & 0.910 [0.881, 0.932] \\
OpenCode (Gemini) & 1.834 [1.592, 2.086] & 0.892 [0.853, 0.921] \\
OpenCode (GLM) & 1.689 [1.505, 1.878] & 0.908 [0.880, 0.930] \\
\midrule
Last-year value & 2.116 [1.815, 2.407] & 0.856 [0.803, 0.897] \\
Ten-year mean & 1.666 [1.490, 1.851] & 0.911 [0.883, 0.931] \\
\bottomrule
\end{tabular}
\end{adjustbox}
\end{table}

\begin{table}[!htbp]
\centering\small
\caption{Carbon emissions: point estimates and 95\% tile-block bootstrap intervals on 220 cases. RMSE is in tonnes C/tile/year.}
\label{tab:appendix-carbon-rmse-r2}
\begin{adjustbox}{max width=\textwidth}
\begin{tabular}{lrr}
\toprule
Method & RMSE $\downarrow$ [95\% CI] & $R^2\uparrow$ [95\% CI] \\
\midrule
Earth Agent (GPT Luna) & 1,043.065 [821.661, 1,212.518] & 0.999 [0.997, 0.999] \\
Earth Agent (DeepSeek) & 989.874 [782.099, 1,160.028] & 0.999 [0.997, 0.999] \\
Codex (GPT-5.5) & 1,013.169 [824.009, 1,163.599] & 0.999 [0.997, 0.999] \\
Codex (DeepSeek) & 1,008.246 [760.608, 1,208.535] & 0.999 [0.997, 0.999] \\
OpenCode (GPT Luna) & 1,052.534 [793.467, 1,264.855] & 0.999 [0.998, 0.999] \\
OpenCode (Gemini) & 1,215.050 [875.505, 1,487.187] & 0.998 [0.997, 0.999] \\
OpenCode (GLM) & 1,050.863 [813.899, 1,251.089] & 0.999 [0.997, 0.999] \\
\midrule
Last-year value & 1,355.533 [1,011.841, 1,593.081] & 0.998 [0.996, 0.998] \\
Ten-year mean & 4,935.183 [3,388.520, 5,968.457] & 0.974 [0.959, 0.980] \\
\bottomrule
\end{tabular}
\end{adjustbox}
\end{table}

\begin{table}[!htbp]
\centering\small
\caption{Property prices: point estimates and 95\% tile-block bootstrap intervals. Each city contains 20 cases from four tiles.}
\label{tab:appendix-property-rmse-r2}
\begin{adjustbox}{max width=\textwidth}
\begin{tabular}{lrr}
\toprule
Method & RMSE $\downarrow$ [95\% CI] & $R^2\uparrow$ [95\% CI] \\
\midrule
\multicolumn{3}{l}{\textit{Boston (RMSE: $10^3$ USD; 4 tiles)}} \\
Earth Agent (GPT Luna) & 28.005 [23.449, 32.922] & 0.959 [0.558, 0.963] \\
Earth Agent (DeepSeek) & 24.255 [22.348, 26.273] & 0.969 [0.657, 0.976] \\
Codex (GPT-5.5) & 22.640 [21.623, 23.764] & 0.973 [0.671, 0.980] \\
Codex (DeepSeek) & 28.421 [20.470, 36.315] & 0.957 [0.535, 0.968] \\
OpenCode (GPT Luna) & 24.800 [21.666, 27.401] & 0.968 [0.542, 0.980] \\
OpenCode (Gemini) & 32.566 [24.606, 42.727] & 0.944 [0.583, 0.948] \\
OpenCode (GLM) & 33.407 [29.426, 36.489] & 0.941 [0.119, 0.962] \\
Last-year value & 34.915 [29.810, 38.153] & 0.936 [0.279, 0.955] \\
Ten-year mean & 189.845 [179.598, 196.409] & -0.901 [-24.491, -0.376] \\
\midrule
\multicolumn{3}{l}{\textit{Busan (RMSE: $10^6$ KRW; 4 tiles)}} \\
Earth Agent (GPT Luna) & 101.851 [29.727, 165.006] & 0.529 [-1.266, 0.912] \\
Earth Agent (DeepSeek) & 96.857 [32.743, 151.730] & 0.574 [-1.058, 0.895] \\
Codex (GPT-5.5) & 105.079 [41.840, 160.921] & 0.499 [-1.485, 0.824] \\
Codex (DeepSeek) & 103.633 [41.209, 159.367] & 0.513 [-1.443, 0.833] \\
OpenCode (GPT Luna) & 108.882 [45.205, 162.141] & 0.462 [-1.598, 0.813] \\
OpenCode (Gemini) & 88.758 [36.586, 133.882] & 0.642 [-0.660, 0.907] \\
OpenCode (GLM) & 89.685 [37.956, 135.411] & 0.635 [-0.756, 0.859] \\
Last-year value & 102.324 [36.132, 159.641] & 0.525 [-1.484, 0.875] \\
Ten-year mean & 104.700 [45.839, 144.383] & 0.502 [-2.282, 0.770] \\
\bottomrule
\end{tabular}
\end{adjustbox}
\end{table}

\begin{table}[!htbp]
\centering\small
\caption{Property prices: point estimates and 95\% tile-block bootstrap intervals. Each city contains 20 cases from four tiles.}
\label{tab:appendix-property-rmse-r2-2}
\begin{adjustbox}{max width=\textwidth}
\begin{tabular}{lrr}
\toprule
Method & RMSE $\downarrow$ [95\% CI] & $R^2\uparrow$ [95\% CI] \\
\midrule
\multicolumn{3}{l}{\textit{DC (RMSE: $10^3$ USD; 4 tiles)}} \\
Earth Agent (GPT Luna) & 37.454 [27.194, 45.455] & 0.982 [-0.537, 0.988] \\
Earth Agent (DeepSeek) & 39.700 [28.299, 48.490] & 0.980 [-0.864, 0.987] \\
Codex (GPT-5.5) & 33.597 [24.462, 40.732] & 0.986 [-0.241, 0.992] \\
Codex (DeepSeek) & 34.626 [24.108, 44.211] & 0.985 [-0.267, 0.989] \\
OpenCode (GPT Luna) & 44.006 [31.672, 51.499] & 0.975 [-1.188, 0.987] \\
OpenCode (Gemini) & 43.856 [33.954, 54.153] & 0.975 [-1.353, 0.981] \\
OpenCode (GLM) & 38.064 [33.614, 42.893] & 0.981 [-1.290, 0.986] \\
Last-year value & 37.626 [23.862, 52.629] & 0.982 [-0.205, 0.986] \\
Ten-year mean & 162.937 [130.264, 207.561] & 0.660 [-33.686, 0.718] \\
\midrule
\multicolumn{3}{l}{\textit{London (RMSE: $10^3$ GBP; 4 tiles)}} \\
Earth Agent (GPT Luna) & 109.601 [45.439, 171.976] & 0.398 [-2.978, 0.750] \\
Earth Agent (DeepSeek) & 107.925 [34.781, 178.928] & 0.416 [-0.823, 0.849] \\
Codex (GPT-5.5) & 107.420 [37.759, 175.376] & 0.422 [-0.695, 0.775] \\
Codex (DeepSeek) & 104.736 [35.205, 170.910] & 0.450 [-0.780, 0.820] \\
OpenCode (GPT Luna) & 102.560 [37.197, 165.496] & 0.473 [-0.886, 0.806] \\
OpenCode (Gemini) & 103.042 [31.021, 164.235] & 0.468 [-0.693, 0.746] \\
OpenCode (GLM) & 104.718 [30.890, 172.017] & 0.451 [-0.642, 0.825] \\
Last-year value & 99.440 [30.641, 163.781] & 0.505 [-0.475, 0.847] \\
Ten-year mean & 115.542 [63.597, 171.475] & 0.331 [-4.073, 0.379] \\
\bottomrule
\end{tabular}
\end{adjustbox}
\end{table}

\begin{table}[!htbp]
\centering\small
\caption{Property prices: point estimates and 95\% tile-block bootstrap intervals. Each city contains 20 cases from four tiles.}
\label{tab:appendix-property-rmse-r2-3}
\begin{adjustbox}{max width=\textwidth}
\begin{tabular}{lrr}
\toprule
Method & RMSE $\downarrow$ [95\% CI] & $R^2\uparrow$ [95\% CI] \\
\midrule
\multicolumn{3}{l}{\textit{New York City (RMSE: $10^3$ USD; 4 tiles)}} \\
Earth Agent (GPT Luna) & 67.079 [40.838, 85.625] & 0.910 [-1.221, 0.947] \\
Earth Agent (DeepSeek) & 64.188 [34.347, 84.027] & 0.917 [-0.793, 0.957] \\
Codex (GPT-5.5) & 61.009 [35.084, 80.067] & 0.925 [-0.691, 0.966] \\
Codex (DeepSeek) & 62.738 [39.748, 75.760] & 0.921 [-1.834, 0.962] \\
OpenCode (GPT Luna) & 72.192 [32.251, 96.866] & 0.895 [-0.857, 0.947] \\
OpenCode (Gemini) & 85.311 [50.303, 111.837] & 0.854 [-2.292, 0.932] \\
OpenCode (GLM) & 65.813 [37.108, 91.010] & 0.913 [-1.123, 0.968] \\
Last-year value & 51.316 [25.703, 67.868] & 0.947 [-0.077, 0.973] \\
Ten-year mean & 170.349 [91.146, 242.217] & 0.417 [-16.862, 0.512] \\
\midrule
\multicolumn{3}{l}{\textit{San Francisco (RMSE: $10^3$ USD; 4 tiles)}} \\
Earth Agent (GPT Luna) & 137.259 [114.793, 156.533] & -0.918 [-3.798, -0.306] \\
Earth Agent (DeepSeek) & 95.075 [83.398, 105.467] & 0.080 [-1.127, 0.300] \\
Codex (GPT-5.5) & 86.821 [79.080, 92.623] & 0.233 [-0.863, 0.484] \\
Codex (DeepSeek) & 82.479 [62.429, 93.559] & 0.308 [-0.975, 0.607] \\
OpenCode (GPT Luna) & 93.658 [74.413, 109.574] & 0.107 [-1.330, 0.336] \\
OpenCode (Gemini) & 128.069 [92.314, 160.319] & -0.669 [-3.596, -0.073] \\
OpenCode (GLM) & 114.857 [97.354, 130.025] & -0.343 [-2.240, 0.086] \\
Last-year value & 77.496 [70.897, 83.576] & 0.389 [-0.492, 0.547] \\
Ten-year mean & 260.979 [195.595, 297.383] & -5.933 [-17.512, -3.266] \\
\bottomrule
\end{tabular}
\end{adjustbox}
\end{table}

\begin{table}[!htbp]
\centering\small
\caption{Property prices: point estimates and 95\% tile-block bootstrap intervals. Each city contains 20 cases from four tiles.}
\label{tab:appendix-property-rmse-r2-4}
\begin{adjustbox}{max width=\textwidth}
\begin{tabular}{lrr}
\toprule
Method & RMSE $\downarrow$ [95\% CI] & $R^2\uparrow$ [95\% CI] \\
\midrule
\multicolumn{3}{l}{\textit{Singapore (RMSE: $10^3$ SGD; 4 tiles)}} \\
Earth Agent (GPT Luna) & 107.225 [52.255, 162.016] & 0.759 [0.355, 0.808] \\
Earth Agent (DeepSeek) & 92.433 [55.224, 127.521] & 0.821 [0.277, 0.847] \\
Codex (GPT-5.5) & 90.460 [43.114, 127.864] & 0.828 [0.561, 0.856] \\
Codex (DeepSeek) & 91.366 [32.303, 134.868] & 0.825 [0.630, 0.857] \\
OpenCode (GPT Luna) & 95.167 [62.199, 127.373] & 0.810 [0.026, 0.845] \\
OpenCode (Gemini) & 97.088 [65.061, 130.549] & 0.802 [0.001, 0.820] \\
OpenCode (GLM) & 94.607 [63.062, 126.453] & 0.812 [0.061, 0.833] \\
Last-year value & 109.267 [50.398, 164.023] & 0.749 [0.400, 0.788] \\
Ten-year mean & 235.602 [112.755, 354.118] & -0.165 [-2.001, 0.047] \\
\midrule
\multicolumn{3}{l}{\textit{Sydney (RMSE: $10^3$ AUD; 4 tiles)}} \\
Earth Agent (GPT Luna) & 231.405 [140.671, 298.131] & 0.876 [-0.411, 0.911] \\
Earth Agent (DeepSeek) & 242.987 [134.795, 346.046] & 0.863 [-0.187, 0.887] \\
Codex (GPT-5.5) & 255.910 [164.569, 318.882] & 0.848 [-0.668, 0.902] \\
Codex (DeepSeek) & 177.384 [106.759, 223.612] & 0.927 [0.161, 0.955] \\
OpenCode (GPT Luna) & 269.539 [165.134, 346.105] & 0.831 [-0.808, 0.884] \\
OpenCode (Gemini) & 241.767 [131.780, 334.735] & 0.864 [-0.106, 0.894] \\
OpenCode (GLM) & 244.571 [158.688, 315.395] & 0.861 [-0.407, 0.899] \\
Last-year value & 263.068 [162.017, 341.523] & 0.839 [-0.573, 0.885] \\
Ten-year mean & 575.934 [301.856, 805.350] & 0.229 [-2.478, 0.425] \\
\bottomrule
\end{tabular}
\end{adjustbox}
\end{table}

\FloatBarrier
\paragraph{Prediction diagnostics.}
Tables~\ref{tab:agent-appendix-lst-groups}--\ref{tab:agent-appendix-property-groups} report errors by city and forecast year. Calendar years are restored for this evaluation breakdown. LST and carbon use MAE in their target units, and property uses MAPE across local currencies. Carbon errors vary substantially with the emission scale of each city. All seven agent configurations use the same evaluation cases. The added Codex (DeepSeek) and OpenCode (GLM) columns complete the city- and year-level comparison.

\begin{table}[!htbp]
\centering\scriptsize
\setlength{\tabcolsep}{3pt}
\caption{Land-surface temperature: MAE by city and forecast year ($^\circ$C).}
\label{tab:agent-appendix-lst-groups}
\resizebox{\textwidth}{!}{%
\begin{tabular}{llrrrrrrrrr}
\toprule
Group & $n$ & Earth--Luna & Earth--DS & OC--Luna & OC--Gemini & Codex--GPT & Codex--DS & OC--GLM & Last year & Mean10 \\
\midrule
\multicolumn{11}{l}{\emph{By city}} \\
Amsterdam & 15 & 1.264 & 1.156 & 1.403 & 1.349 & 1.177 & 1.302 & 1.322 & 1.705 & 1.198 \\
Boston & 15 & 1.269 & 1.403 & 1.309 & 1.290 & 1.223 & 1.326 & 1.341 & 2.063 & 1.216 \\
Busan & 15 & 0.723 & 0.675 & 0.825 & 0.759 & 0.674 & 0.663 & 0.803 & 0.836 & 0.561 \\
Cape Town & 15 & 1.264 & 1.209 & 1.291 & 1.193 & 1.154 & 1.235 & 1.374 & 1.049 & 1.602 \\
Christchurch & 15 & 1.387 & 1.207 & 1.220 & 1.428 & 1.173 & 1.254 & 1.370 & 1.355 & 1.270 \\
DC & 15 & 1.204 & 1.254 & 1.096 & 1.193 & 1.055 & 1.398 & 1.034 & 1.720 & 1.037 \\
London & 15 & 2.087 & 2.308 & 2.020 & 2.354 & 1.952 & 1.821 & 1.971 & 2.746 & 1.985 \\
Lyon & 15 & 2.123 & 1.927 & 1.996 & 2.061 & 2.123 & 2.019 & 1.985 & 3.047 & 1.885 \\
New York City & 15 & 1.285 & 1.153 & 1.163 & 1.373 & 1.195 & 1.162 & 1.238 & 1.973 & 1.084 \\
San Francisco & 15 & 1.418 & 1.393 & 1.254 & 1.623 & 1.383 & 1.137 & 1.175 & 1.325 & 1.087 \\
Santiago & 15 & 1.669 & 1.723 & 1.386 & 1.819 & 1.625 & 1.459 & 1.551 & 1.472 & 1.359 \\
Singapore & 10 & 0.904 & 0.850 & 0.825 & 0.931 & 0.862 & 0.888 & 0.898 & 0.848 & 0.640 \\
Sydney & 15 & 1.829 & 1.611 & 1.609 & 1.684 & 1.478 & 1.703 & 1.428 & 1.564 & 1.551 \\
Tokyo & 15 & 1.045 & 1.275 & 1.172 & 1.128 & 1.150 & 0.999 & 1.208 & 1.414 & 1.134 \\
Xiong'an & 15 & 1.886 & 1.679 & 1.573 & 1.691 & 1.728 & 1.724 & 1.566 & 1.766 & 1.863 \\
\midrule
\multicolumn{11}{l}{\emph{By forecast year}} \\
2020 & 44 & 1.256 & 1.368 & 1.231 & 1.292 & 1.218 & 1.214 & 1.337 & 1.688 & 1.240 \\
2021 & 44 & 1.776 & 1.538 & 1.587 & 1.734 & 1.676 & 1.346 & 1.537 & 1.949 & 1.437 \\
2022 & 44 & 1.403 & 1.326 & 1.192 & 1.323 & 1.225 & 1.428 & 1.199 & 1.608 & 1.062 \\
2023 & 44 & 1.355 & 1.438 & 1.384 & 1.535 & 1.299 & 1.380 & 1.378 & 1.478 & 1.380 \\
2024 & 44 & 1.390 & 1.333 & 1.378 & 1.467 & 1.287 & 1.379 & 1.355 & 1.664 & 1.447 \\
\bottomrule
\end{tabular}}
\par\smallskip\footnotesize DS: DeepSeek; Codex--GPT: Codex (GPT-5.5); OC: OpenCode.
\end{table}

\begin{table}[!htbp]
\centering\scriptsize
\setlength{\tabcolsep}{3pt}
\caption{Carbon emissions: MAE by city and forecast year (tonnes C/tile/year).}
\label{tab:agent-appendix-carbon-groups}
\resizebox{\textwidth}{!}{%
\begin{tabular}{llrrrrrrrrr}
\toprule
Group & $n$ & Earth--Luna & Earth--DS & OC--Luna & OC--Gemini & Codex--GPT & Codex--DS & OC--GLM & Last year & Mean10 \\
\midrule
\multicolumn{11}{l}{\emph{By city}} \\
Amsterdam & 15 & 276.613 & 226.749 & 288.957 & 281.078 & 295.122 & 287.385 & 305.042 & 297.730 & 1,006.523 \\
Boston & 10 & 114.388 & 132.012 & 121.332 & 107.740 & 93.660 & 36.772 & 84.891 & 123.812 & 180.381 \\
Busan & 15 & 514.681 & 488.062 & 581.498 & 517.622 & 473.480 & 287.836 & 453.674 & 304.751 & 403.713 \\
Cape Town & 15 & 331.489 & 331.532 & 281.313 & 288.556 & 380.377 & 324.304 & 318.488 & 256.331 & 385.654 \\
Christchurch & 15 & 0.084 & 0.073 & 0.074 & 0.064 & 0.068 & 0.047 & 0.088 & 0.061 & 0.073 \\
DC & 15 & 260.834 & 291.934 & 335.708 & 295.742 & 309.908 & 207.082 & 289.724 & 366.783 & 504.131 \\
London & 15 & 194.675 & 151.211 & 154.035 & 214.126 & 149.211 & 89.947 & 109.989 & 159.943 & 801.164 \\
Lyon & 15 & 379.862 & 438.242 & 385.666 & 375.469 & 505.938 & 216.276 & 379.548 & 529.565 & 1,149.527 \\
New York City & 15 & 276.473 & 345.825 & 335.307 & 327.485 & 306.492 & 222.601 & 361.929 & 352.280 & 513.236 \\
San Francisco & 15 & 38.262 & 41.215 & 48.304 & 41.091 & 37.875 & 27.891 & 46.722 & 49.863 & 72.645 \\
Santiago & 15 & 1,867.614 & 1,465.229 & 1,509.240 & 1,727.339 & 1,424.114 & 1,649.269 & 1,500.225 & 1,377.077 & 824.526 \\
Shanghai & 15 & 1,871.851 & 1,380.796 & 1,905.988 & 2,218.115 & 1,760.247 & 1,820.074 & 1,931.189 & 3,189.553 & 13,648.485 \\
Singapore & 15 & 909.223 & 1,184.343 & 948.862 & 932.095 & 1,032.982 & 1,176.462 & 1,106.407 & 1,562.649 & 8,494.535 \\
Sydney & 10 & 609.744 & 511.565 & 529.353 & 728.179 & 480.664 & 587.984 & 647.427 & 314.393 & 674.577 \\
Tokyo & 10 & 861.923 & 999.746 & 877.195 & 1,091.602 & 1,248.750 & 490.768 & 1,080.500 & 929.696 & 4,356.582 \\
Xiong'an & 10 & 100.146 & 94.542 & 68.852 & 104.682 & 91.841 & 107.798 & 140.677 & 187.160 & 800.882 \\
\midrule
\multicolumn{11}{l}{\emph{By forecast year}} \\
2020 & 44 & 873.673 & 701.876 & 737.474 & 815.289 & 888.710 & 645.999 & 845.825 & 967.304 & 2,113.089 \\
2021 & 44 & 469.740 & 502.044 & 517.421 & 435.300 & 410.509 & 411.281 & 454.962 & 467.062 & 1,816.894 \\
2022 & 44 & 386.049 & 373.406 & 337.557 & 300.538 & 385.077 & 338.271 & 357.558 & 556.525 & 1,984.669 \\
2023 & 44 & 628.651 & 681.511 & 539.449 & 819.490 & 653.820 & 569.511 & 602.242 & 944.726 & 2,524.329 \\
2024 & 44 & 384.771 & 299.272 & 540.634 & 552.196 & 372.937 & 463.821 & 502.602 & 297.324 & 2,406.187 \\
\bottomrule
\end{tabular}}
\par\smallskip\footnotesize DS: DeepSeek; Codex--GPT: Codex (GPT-5.5); OC: OpenCode.
\end{table}

\begin{table}[!htbp]
\centering\scriptsize
\setlength{\tabcolsep}{3pt}
\caption{Property prices: MAPE by city and forecast year (\%).}
\label{tab:agent-appendix-property-groups}
\resizebox{\textwidth}{!}{%
\begin{tabular}{llrrrrrrrrr}
\toprule
Group & $n$ & Earth--Luna & Earth--DS & OC--Luna & OC--Gemini & Codex--GPT & Codex--DS & OC--GLM & Last year & Mean10 \\
\midrule
\multicolumn{11}{l}{\emph{By city}} \\
Boston & 20 & 3.120 & 2.897 & 3.137 & 3.768 & 2.887 & 3.278 & 4.152 & 4.460 & 28.308 \\
Busan & 20 & 15.627 & 17.512 & 19.443 & 15.539 & 17.520 & 19.124 & 15.304 & 17.269 & 18.847 \\
DC & 20 & 5.112 & 5.548 & 5.996 & 6.237 & 4.577 & 4.699 & 5.370 & 4.803 & 27.320 \\
London & 20 & 14.454 & 11.929 & 12.367 & 12.114 & 12.240 & 12.274 & 12.143 & 10.916 & 18.306 \\
New York City & 20 & 6.237 & 5.645 & 6.221 & 8.055 & 5.286 & 5.265 & 5.916 & 3.987 & 16.074 \\
San Francisco & 20 & 9.442 & 5.780 & 5.888 & 8.702 & 4.990 & 4.923 & 7.633 & 5.159 & 18.916 \\
Singapore & 20 & 11.074 & 9.794 & 10.809 & 11.725 & 9.022 & 7.927 & 10.656 & 10.963 & 23.436 \\
Sydney & 20 & 10.709 & 10.014 & 11.545 & 10.597 & 11.256 & 8.214 & 10.841 & 11.767 & 28.059 \\
\midrule
\multicolumn{11}{l}{\emph{By forecast year}} \\
2020 & 32 & 6.991 & 6.405 & 6.674 & 6.595 & 5.792 & 6.084 & 6.040 & 7.037 & 22.037 \\
2021 & 32 & 10.328 & 9.857 & 11.311 & 10.856 & 9.507 & 8.977 & 10.928 & 11.427 & 25.402 \\
2022 & 32 & 8.074 & 8.377 & 9.119 & 9.274 & 8.291 & 8.592 & 9.333 & 7.976 & 25.017 \\
2023 & 32 & 11.673 & 9.934 & 10.761 & 10.664 & 10.935 & 10.160 & 10.025 & 8.895 & 19.635 \\
2024 & 32 & 10.293 & 8.626 & 9.263 & 10.571 & 7.837 & 7.253 & 8.682 & 7.993 & 19.949 \\
\bottomrule
\end{tabular}}
\par\smallskip\footnotesize DS: DeepSeek; Codex--GPT: Codex (GPT-5.5); OC: OpenCode.
\end{table}

\FloatBarrier
\paragraph{Planning diagnostics.}
Table~\ref{tab:agent-appendix-ev-confusion} shows a common concentration of EV predictions in the lower expansion classes. Across the six configurations, 6--9 of the 37 class-D cases are correctly identified. Codex (DeepSeek) recovers 9 class-D cases and assigns 28 to lower expansion ranges. Table~\ref{tab:agent-appendix-school-plans} shows variation in seat-addition policies. Codex (DeepSeek) proposes a median of 11 additional seats, retains existing capacity in 45\% of cases, and achieves a balance rate of 89\%. Its remaining cases comprise eight shortages and three surpluses. These patterns connect enrollment forecasts and expansion margins to the final balance outcome.

\begin{table}[!htbp]
\centering
\footnotesize
\setlength{\tabcolsep}{3pt}
\caption{EV charging-port growth: reference-by-prediction confusion counts on the 150-case evaluation cohort. Recall is the diagonal count divided by the reference-row total. A--D denote the city-specific growth ranges.}
\label{tab:agent-appendix-ev-confusion}
\begin{tabular}{llrrrrr}
\toprule
Agent & Reference & A & B & C & D & Recall (\%) \\
\midrule
Earth--Terra & A & 19 & 14 & 5 & 0 & 50.000 \\
 & B & 13 & 20 & 5 & 0 & 52.632 \\
 & C & 12 & 17 & 5 & 3 & 13.514 \\
 & D & 4 & 15 & 9 & 9 & 24.324 \\
\midrule
Earth--DeepSeek & A & 27 & 11 & 0 & 0 & 71.053 \\
 & B & 8 & 23 & 5 & 2 & 60.526 \\
 & C & 13 & 17 & 4 & 3 & 10.811 \\
 & D & 9 & 12 & 7 & 9 & 24.324 \\
\midrule
Codex--GPT-5.5 & A & 20 & 17 & 1 & 0 & 52.632 \\
 & B & 3 & 27 & 8 & 0 & 71.053 \\
 & C & 7 & 22 & 6 & 2 & 16.216 \\
 & D & 3 & 17 & 9 & 8 & 21.622 \\
\midrule
OpenCode--Terra & A & 28 & 8 & 2 & 0 & 73.684 \\
 & B & 20 & 12 & 5 & 1 & 31.579 \\
 & C & 16 & 12 & 6 & 3 & 16.216 \\
 & D & 9 & 13 & 7 & 8 & 21.622 \\
\midrule
OpenCode--Gemini & A & 31 & 7 & 0 & 0 & 81.579 \\
 & B & 25 & 10 & 3 & 0 & 26.316 \\
 & C & 25 & 7 & 3 & 2 & 8.108 \\
 & D & 19 & 9 & 3 & 6 & 16.216 \\
\midrule
Codex--DeepSeek & A & 23 & 15 & 0 & 0 & 60.526 \\
 & B & 8 & 18 & 11 & 1 & 47.368 \\
 & C & 13 & 14 & 7 & 3 & 18.919 \\
 & D & 8 & 11 & 9 & 9 & 24.324 \\
\bottomrule
\end{tabular}
\end{table}

\begin{table}[!htbp]
\centering
\footnotesize
\setlength{\tabcolsep}{3pt}
\caption{School capacity plans across 100 cases. Planned capacity is the last observed capacity plus proposed seats; balance requires observed next-period enrollment to be 95\%--105\% of planned capacity.}
\label{tab:agent-appendix-school-plans}
\begin{tabular}{lrrrrr}
\toprule
Agent & Added seats: median [IQR] & Zero addition (\%) & Balanced (\%) & Shortage (\%) & Surplus (\%) \\
\midrule
Earth--Terra & 0.000 [0.000, 38.250] & 57.000 & 81.000 & 16.000 & 3.000 \\
Earth--DeepSeek & 20.000 [0.000, 75.750] & 35.000 & 88.000 & 7.000 & 5.000 \\
Codex--GPT-5.5 & 0.000 [0.000, 46.500] & 57.000 & 84.000 & 13.000 & 3.000 \\
OpenCode--Terra & 0.000 [0.000, 33.750] & 53.000 & 80.000 & 17.000 & 3.000 \\
OpenCode--Gemini & 16.500 [0.000, 61.500] & 38.000 & 84.000 & 10.000 & 6.000 \\
Codex--DeepSeek & 11.000 [0.000, 67.000] & 45.000 & 89.000 & 8.000 & 3.000 \\
\bottomrule
\end{tabular}
\end{table}

\FloatBarrier
\paragraph{Resource-analysis diagnostics.}
Tables~\ref{tab:agent-appendix-food-confusion} and~\ref{tab:agent-appendix-food-spatial} describe class-specific errors and within-tile classification. Codex (DeepSeek) correctly identifies 28 of the 62 low-income-and-low-access cells, giving 45.161\% recall. Its cell accuracy reaches 51.316\% on mixed-label tiles, while complete four-cell agreement is achieved on 4 of 57 such tiles (7.018\%). Across the six configurations, mixed-label exact-tile accuracy ranges from 0 to 7.018\%. The results identify the recovery of local spatial differences as a persistent challenge alongside class recognition.

\begin{table}[!htbp]
\centering
\small
\setlength{\tabcolsep}{4pt}
\caption{Food-access cell-label confusion counts on the 150-tile evaluation cohort. NLA: not low access; LA: low access; LILA: low income and low access. Rows are reference classes and columns are predicted classes.}
\label{tab:agent-appendix-food-confusion}
\begin{minipage}[t]{0.49\linewidth}
\vspace{0pt}\centering\footnotesize
\resizebox{\linewidth}{!}{%
\begin{tabular}{@{}llrrrr@{}}
\toprule
Agent & Reference & NLA & LA & LILA & Recall (\%) \\
\midrule
Earth--Terra & NLA & 242 & 22 & 34 & 81.208 \\
 & LA & 188 & 43 & 9 & 17.917 \\
 & LILA & 38 & 9 & 15 & 24.194 \\
\midrule
Earth--DeepSeek & NLA & 224 & 68 & 6 & 75.168 \\
 & LA & 146 & 78 & 16 & 32.500 \\
 & LILA & 45 & 11 & 6 & 9.677 \\
\midrule
Codex--GPT-5.5 & NLA & 218 & 29 & 51 & 73.154 \\
 & LA & 143 & 75 & 22 & 31.250 \\
 & LILA & 13 & 19 & 30 & 48.387 \\
\bottomrule
\end{tabular}}
\end{minipage}\hfill
\begin{minipage}[t]{0.49\linewidth}
\vspace{0pt}\centering\footnotesize
\resizebox{\linewidth}{!}{%
\begin{tabular}{@{}llrrrr@{}}
\toprule
Agent & Reference & NLA & LA & LILA & Recall (\%) \\
\midrule
OpenCode--Terra & NLA & 259 & 14 & 25 & 86.913 \\
 & LA & 186 & 43 & 11 & 17.917 \\
 & LILA & 31 & 11 & 20 & 32.258 \\
\midrule
OpenCode--Gemini & NLA & 219 & 74 & 5 & 73.490 \\
 & LA & 116 & 121 & 3 & 50.417 \\
 & LILA & 35 & 25 & 2 & 3.226 \\
\midrule
Codex--DeepSeek & NLA & 211 & 49 & 38 & 70.805 \\
 & LA & 128 & 93 & 19 & 38.750 \\
 & LILA & 23 & 11 & 28 & 45.161 \\
\bottomrule
\end{tabular}}
\end{minipage}
\end{table}

\begin{table}[!htbp]
\centering
\small
\setlength{\tabcolsep}{5pt}
\caption{Food-access performance by within-tile reference-label composition: 93 uniform-label tiles and 57 mixed-label tiles. Cell accuracy counts correct labels across all four cells; exact-tile accuracy requires all four to be correct.}
\label{tab:agent-appendix-food-spatial}
\begin{tabular}{lrrrr}
\toprule
Agent & Uniform: cell (\%) & Uniform: exact tile (\%) & Mixed: cell (\%) & Mixed: exact tile (\%) \\
\midrule
Earth--Terra & 56.183 & 53.763 & 39.912 & 0.000 \\
Earth--DeepSeek & 52.957 & 21.505 & 48.684 & 3.509 \\
Codex--GPT-5.5 & 61.559 & 58.065 & 41.228 & 3.509 \\
OpenCode--Terra & 57.527 & 53.763 & 47.368 & 5.263 \\
OpenCode--Gemini & 68.548 & 61.290 & 38.158 & 1.754 \\
Codex--DeepSeek & 57.796 & 25.806 & 51.316 & 7.018 \\
\bottomrule
\end{tabular}
\end{table}

\FloatBarrier
\subsubsection{Tool Use and Task Performance}
\label{app:agent-tool-use}

\paragraph{Comparison protocol.}
Tables~\ref{tab:forecast-tool-use}, \ref{tab:planning-resource-tool-use}, and~\ref{tab:food-tool-use} pair each configuration's average tool-call count with its final task performance on the same evaluation cases. Calls are counted in the attempt selected for scoring, using the first successful submission when a case has retries. Each native invocation is counted once, including calls that return errors. Submission-tool invocations and automatic answer processing are excluded. Forecasting counts native shell and file operations as well as task-specific analytical tools; planning and resource analysis count the shared task-specific tools. A shell invocation can contain multiple reads and calculations. The tables therefore compare interaction volume alongside the task's evaluation metrics. Bold values indicate the best task performance among the agent configurations in each column.

\paragraph{Forecasting.}
The largest call count coincides with the best carbon and property forecasts for Codex (DeepSeek). LST shows a different pattern: Codex (GPT-5.5) achieves lower MAE with 17.068 calls per case, compared with 46.982 for Codex (DeepSeek). OpenCode (GPT Luna) also makes more LST calls than Codex (GPT-5.5) and has higher error. Across forecasting targets, the value of additional tool use depends on the analysis and numerical forecasting rule selected from the retrieved evidence.

\begin{table}[!htbp]
\centering\small
\setlength{\tabcolsep}{4pt}
\renewcommand{\arraystretch}{1.10}
\caption{Tool-call volume and forecasting performance. Calls are mean invocations per case; each task pairs call volume with its final forecast error.}
\label{tab:forecast-tool-use}
\begin{adjustbox}{max width=\textwidth}
\begin{tabular}{lrrrrrr}
\toprule
& \multicolumn{2}{c}{LST ($n=220$)} & \multicolumn{2}{c}{Carbon ($n=220$)} & \multicolumn{2}{c}{Property ($n=160$)} \\
\cmidrule(lr){2-3}\cmidrule(lr){4-5}\cmidrule(lr){6-7}
Agent & Calls & MAE $\downarrow$ & Calls & MAE $\downarrow$ & Calls & MAPE (\%) $\downarrow$ \\
\midrule
Earth Agent (GPT Luna) & 18.850 & 1.436 & 11.405 & 548.58 & 11.738 & 9.47 \\
Earth Agent (DeepSeek) & 27.491 & 1.400 & 20.350 & 511.62 & 17.881 & 8.64 \\
Codex (GPT-5.5) & 17.068 & \textbf{1.341} & 17.155 & 542.21 & 16.269 & 8.47 \\
Codex (DeepSeek) & 46.982 & 1.350 & 43.368 & \textbf{485.78} & 45.644 & \textbf{8.21} \\
OpenCode (GPT Luna) & 26.186 & 1.354 & 24.618 & 534.51 & 24.556 & 9.43 \\
OpenCode (Gemini) & 11.673 & 1.470 & 11.205 & 584.56 & 11.262 & 9.59 \\
OpenCode (GLM) & 11.209 & 1.361 & 10.900 & 552.64 & 10.750 & 9.00 \\
\bottomrule
\end{tabular}
\end{adjustbox}
\par\smallskip
\begin{minipage}{\textwidth}\footnotesize
LST MAE is in $^\circ$C; carbon MAE is in tonnes C/tile/year. Property uses MAPE across local currencies.
\end{minipage}
\end{table}

\FloatBarrier
\paragraph{Planning.}
In EV charging, Codex (DeepSeek) makes the most calls, while Earth Agent (DeepSeek) achieves the highest accuracy and Codex (GPT-5.5) the highest macro-F1. In school planning, Codex (DeepSeek) achieves the best balance rate and mean balance with fewer calls than Earth Agent (DeepSeek).

\begin{table}[!htbp]
\centering\small
\setlength{\tabcolsep}{4pt}
\renewcommand{\arraystretch}{1.10}
\caption{Tool-call volume and planning performance. Calls are mean invocations per case.}
\label{tab:planning-resource-tool-use}
\begin{adjustbox}{max width=\textwidth}
\begin{tabular}{lrrrrrr}
\toprule
& \multicolumn{3}{c}{EV charging ($n=150$)} & \multicolumn{3}{c}{School capacity ($n=100$)} \\
\cmidrule(lr){2-4}\cmidrule(lr){5-7}
Agent & Calls & Acc. (\%) $\uparrow$ & Macro-F1 $\uparrow$ & Calls & Balance (\%) $\uparrow$ & Mean balance $\uparrow$ \\
\midrule
Earth Agent (GPT Terra) & 23.860 & 35.3 & 0.339 & 1.330 & 81.0 & 96.91 \\
Earth Agent (DeepSeek) & 20.913 & \textbf{42.0} & 0.382 & 8.020 & 88.0 & 97.50 \\
Codex (GPT-5.5) & 11.273 & 40.7 & \textbf{0.387} & 1.770 & 84.0 & 97.20 \\
Codex (DeepSeek) & 25.833 & 38.0 & 0.366 & 6.990 & \textbf{89.0} & \textbf{97.61} \\
OpenCode (GPT Terra) & 25.067 & 36.0 & 0.333 & 2.650 & 80.0 & 96.74 \\
OpenCode (Gemini) & 4.793 & 33.3 & 0.282 & 1.160 & 84.0 & 97.45 \\
\bottomrule
\end{tabular}
\end{adjustbox}
\par\smallskip
\begin{minipage}{\textwidth}\footnotesize
School balance rate is the percentage of cases with observed enrollment-to-planned-capacity utilization in the 95--105\% range.
\end{minipage}
\end{table}

\FloatBarrier
\paragraph{Resource analysis.}
Food access shows a trade-off across evaluation metrics: Codex (DeepSeek) combines the highest call count with the highest macro-F1, while OpenCode (Gemini) achieves higher cell and exact-case accuracy with substantially fewer calls.

\begin{table}[!htbp]
\centering\small
\setlength{\tabcolsep}{4pt}
\renewcommand{\arraystretch}{1.10}
\caption{Tool-call volume and food-access performance on 150 cases (600 cells). Calls are mean invocations per case.}
\label{tab:food-tool-use}
\begin{adjustbox}{max width=\textwidth}
\begin{tabular}{lrrrr}
\toprule
Agent & Calls & Cell acc. (\%) $\uparrow$ & Macro-F1 $\uparrow$ & Exact-case acc. (\%) $\uparrow$ \\
\midrule
Earth Agent (GPT Terra) & 16.360 & 50.00 & 0.385 & 33.3 \\
Earth Agent (DeepSeek) & 28.520 & 51.33 & 0.385 & 14.7 \\
Codex (GPT-5.5) & 12.907 & 53.83 & 0.475 & 37.3 \\
Codex (DeepSeek) & 49.220 & 55.33 & \textbf{0.498} & 18.7 \\
OpenCode (GPT Terra) & 18.640 & 53.67 & 0.429 & 35.3 \\
OpenCode (Gemini) & 10.933 & \textbf{57.00} & 0.412 & \textbf{38.7} \\
\bottomrule
\end{tabular}
\end{adjustbox}
\par\smallskip
\begin{minipage}{\textwidth}\footnotesize
Exact-case accuracy requires all four cell labels in a tile to be correct.
\end{minipage}
\end{table}

\FloatBarrier
Across tasks, configurations with similar or lower call counts often achieve stronger results, while the most active configuration leads on selected metrics. The case studies identify historical-window selection, capacity margins, and interpretation of spatial evidence as concrete points where retrieved information shapes the final decision.

\FloatBarrier
\subsubsection{Modality Access in Forecasting}
\label{app:forecast-modality-access}
We audit modality access in the selected successful execution trace for each of the 600 Codex (DeepSeek) forecasting cases. Each case contributes at most one count to each modality. A recorded read of an image or raster file counts as image/raster access, including pixel values, numerical summaries, file headers, and metadata. Table-only access denotes cases that read tabular data for that modality and have no recorded image/raster read for it. Modalities corresponding to the forecasting target are included. Access rates use all cases of the corresponding task as the denominator: 220 for LST, 220 for carbon, and 160 for property. Data availability and bulk-reading commands both influence these rates.

The task-specific access patterns emphasize different urban signals. In LST forecasting, weather rasters and optical imagery are accessed in 66.8\% and 66.4\% of cases, respectively. These sources describe climatic conditions and surface characteristics; vegetation and built-form information add context about cooling and heat storage. For carbon emissions, energy-use rasters are accessed in 52.7\% of cases, compared with 25.9\% for LST and 27.5\% for property. Energy use provides activity-related evidence, while nighttime lights and population offer proxies whose interpretation depends on energy intensity and the composition of emission sources. In property forecasting, population, GDP/employment, and built-volume data are accessed in 99.4\%, 98.1\%, and 97.5\% of cases, covering potential demand and supply signals. These patterns show task-dependent evidence selection. The case studies further identify how that evidence enters the numerical decision.

\begin{table}[!htbp]
\centering
\footnotesize
\caption{Forecasting performance grouped by recorded image/raster access for Codex (DeepSeek). The two access groups partition each task's evaluation cases.}
\label{tab:forecast-image-access}
\begin{tabular}{llrrr}
\toprule
Task & Image/raster access & $n$ & MAE $\downarrow$ & $R^2\uparrow$ \\
\midrule
LST & Yes & 198 & 1.359 & 0.9022 \\
LST & No & 22 & 1.268 & 0.9391 \\
Carbon & Yes & 203 & 486.841 & 0.9989 \\
Carbon & No & 17 & 473.070 & 0.9977 \\
\midrule
\multicolumn{3}{l}{Property: image/raster access, $n=147$} & \multicolumn{2}{r}{MAPE = 8.472\%} \\
\multicolumn{3}{l}{Property: no image/raster access, $n=13$} & \multicolumn{2}{r}{MAPE = 5.282\%} \\
\bottomrule
\end{tabular}
\end{table}

\FloatBarrier
The groups differ in case count, city composition, and target-value distribution. Their metrics summarize the forecasts produced on those observed subsets. The image/raster-access groups have higher mean error in all three tasks, while carbon $R^2$ is higher in the access group.

\end{document}